\documentclass{article}

\title{Top-Tuning: a study on transfer learning for an efficient alternative to fine tuning for image classification with fast kernel methods}

\author{%
	{\bf Paolo Didier Alfano}\hfill \hspace{11em}{\small\texttt{paolo.alfano@iit.it}}\\
	{\small \it MaLGa - DIBRIS, University of Genova, Italy \hfill \hspace{12em}}\\
	{\small \it Istituto Italiano di Tecnologia, Genova, Italy \hfill \hspace{12em}}
	\and
	{\bf Vito Paolo Pastore} \hfill \hspace{8.1em}{\small\texttt{vito.paolo.pastore@unige.it}}\\
	{\small \it MaLGa - DIBRIS, University of Genova, Italy \hfill \hspace{12em}}
	\and  
	{\bf Lorenzo Rosasco} \hfill \hspace{10.3em}{\small\texttt{lorenzo.rosasco@unige.it}}\\
	{\small \it MaLGa - DIBRIS, University of Genova, Italy \hfill \hspace{12em}}\\
	{\small \it Istituto Italiano di Tecnologia, Genova, Italy \hfill \hspace{12em}}\\
	{\small \it CBMM - MIT, Cambridge, MA, USA \hfill \hspace{12em}}
	\and
	{\bf Francesca Odone} \hfill \hspace{10.3em}{\small\texttt{francesca.odone@unige.it}}\\
	{\small \it MaLGa - DIBRIS, University of Genova, Italy \hfill \hspace{12em}}
}

\usepackage{bm}
\usepackage{amsfonts}
\usepackage{array,multirow,graphicx}
\usepackage{hyperref}
\usepackage{amsmath}
\usepackage{diagbox}
\date{}

\begin{document}
	\bibliographystyle{plain}
	
	\maketitle
	\begin{abstract}
		The impressive performance of deep learning architectures is associated with a massive increase in model complexity. Millions of parameters need to be tuned, with training and inference time scaling accordingly, together with energy consumption. But is massive fine-tuning always necessary? 
		In this paper, focusing on image classification, we consider a simple transfer learning approach exploiting pre-trained convolutional features as input for a fast-to-train kernel method. We refer to this approach as \textit{top-tuning} since only the kernel classifier is trained on the target dataset. In our study, we perform more than 3000 training processes focusing on 32 small to medium-sized target datasets, a typical situation where transfer learning is necessary. We show that the top-tuning approach provides comparable accuracy with respect to fine-tuning, with a training time between one and two orders of magnitude smaller.
		These results suggest that top-tuning is an effective alternative to fine-tuning in small/medium datasets, being especially useful when training time efficiency and computational resources saving are crucial.

	\end{abstract}
	
	\section{Introduction}
	
	In the last decade, deep learning has led to unprecedented successes in computer vision, at par with human performances in several tasks\cite{NIPS2012_4824, voulodimos2018deep, o2019deep}. In particular, Convolutional Neural Networks (CNNs)\cite{gu2018recent, rawat2017deep} proved successful in a wide range of domains\cite{li2021survey},  from medical images\cite{anwar2018medical, yamashita2018convolutional, MORO2022107119} to robotics \cite{allard2016convolutional, kumra2017robotic} and cyber-security\cite{zheng2017wide, li2017intrusion}, to name a few examples. These advances are related to a frenetic increase in model complexity\cite{szegedy2015going, han2017deep, vaswani2017attention}, need for data\cite{gu2018recent} and corresponding growth of computations\cite{canziani2016analysis}  and energy consumption \cite{strubell2019energy}. The trend is in contrast with an urgent need to set a resources budget, whenever a budget is needed or a more sustainable approach is desired.
	
	Limited data availability is typical of several applications, thus many strategies have been proposed to alleviate the need for labeled data, from few-shot learning\cite{snell2017prototypical, wang2020generalizing} to self-supervision techniques\cite{doersch2017multi, zhai2019s4l}, and also to reduce inference time, e.g. via pruning techniques \cite{248452, DBLP:journals/corr/LiKDSG16}. 
	A budget in computational resources during the training process appears to be less common, as we are observing a visible tendency towards deeper and wider models, whereas training from scratch can be computationally prohibitive
	\cite{7723730, strubell2019energy}.
	
\begin{figure*}[!t]
		\centering
		{\includegraphics[width=0.9\textwidth]{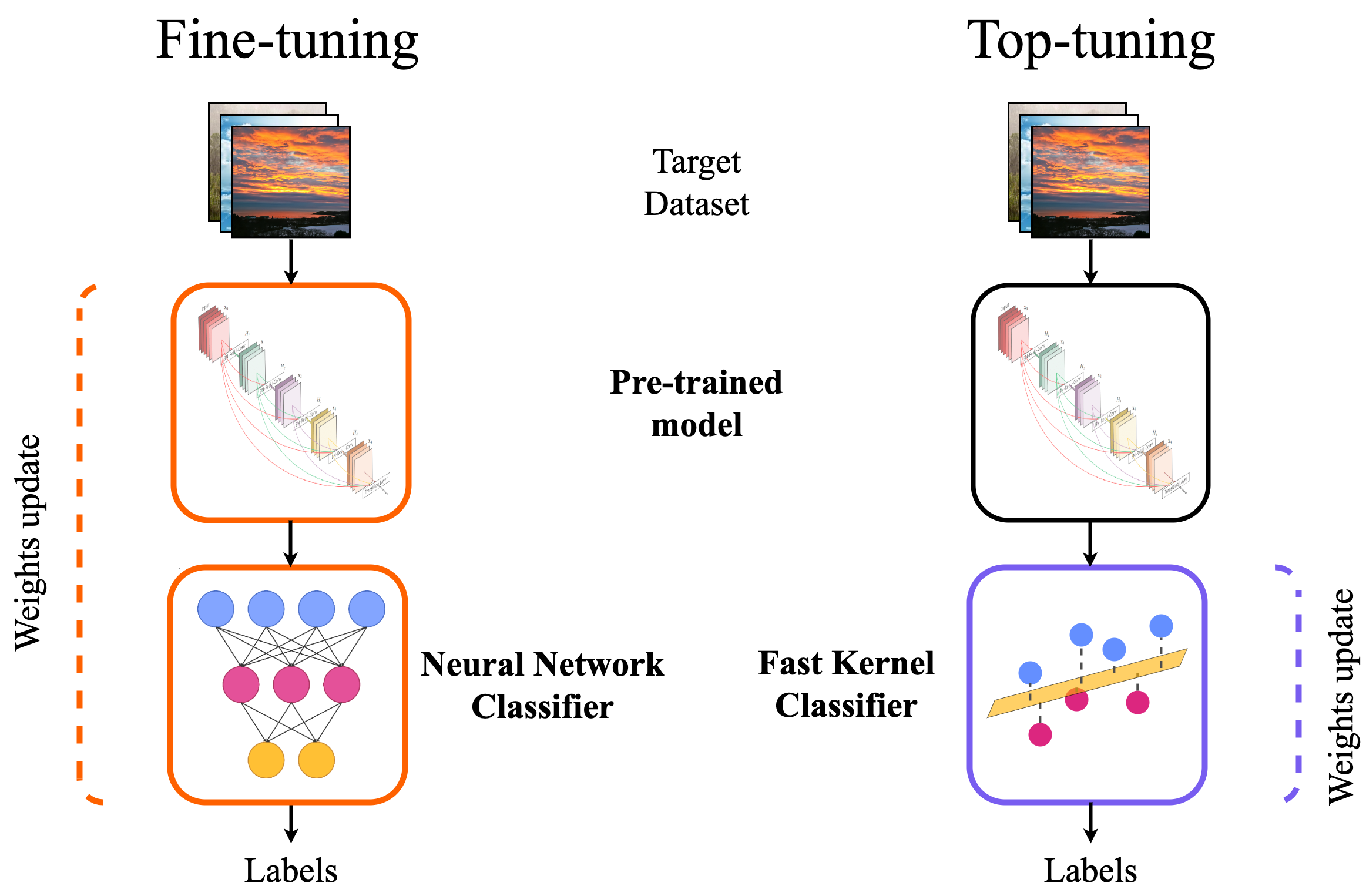}}
		\hspace{3mm}
		\caption{The two pipelines. (Left) The \textit{Fine-Tuning} pipeline. All the model weights are updated. (Right) The \textit{Top-tuning} pipeline. Only the fast kernel weights are updated.}
		\label{fig:Pipeline}
	\end{figure*}

	Transfer learning\cite{pan2009survey, zhuang2020comprehensive}, which aims at transferring knowledge across different source domains, can help tackle both the issues of data scarcity\cite{DBLP:journals/corr/abs-1711-05099, zhuang2020comprehensive} and long training times\cite{oztel2019performance, zhuang2020comprehensive}.
	%, by leveraging pre-trained models to address new problems. 
	The knowledge learned on a problem is stored in the weights of the model. In many practical scenarios with limited availability of data and computational resources, these pre-trained weights represent a good starting configuration to obtain a more refined knowledge of the new task. In particular, a very effective approach is the so-called \textit{fine-tuning}\cite{ZHANG2018146, li2021survey} strategy, where the weights of the network are initialized with pre-trained models and only (fine) tuned rather than being trained from scratch. This approach is shown in \autoref{fig:Pipeline}(Left).\\
	
	Instead, in this paper we propose an approach whose goal is to exploit efficient kernel-based algorithms that are seldom applied to image classification: to compensate for their lack of representation capability, we adopt convolutional features\cite{sharif2014cnn} produced by a state-of-the-art model pre-trained on ImageNet \cite{deng2009imagenet}, with no further tuning. These features are used as input to train a fast and scalable kernel classifier anew\cite{rudi2018falkon, meanti2020kernel}. This is a simple transfer approach\cite{DBLP:journals/corr/Garcia-GasullaP17, 9012507} aiming at exploiting knowledge from the pre-train source. We re-examine such methodology in terms of common practices for transfer learning combining deep nets and fast and scalable kernel methods.\\
	We show that the adopted model, based on Nystr{\"o}m approach to reduce the problem size and on a preconditioned gradient solver for kernel methods, has a remarkable impact on training time and scalability. We refer to such approach as \textit{top-tuning} and we show it in \autoref{fig:Pipeline}(Right).
	
	To investigate the common transfer learning scenario we focus on small to medium-size dataset, where transfer learning is more commonly applied. %We did not include large datasets in our analysis since the transfer learning technique is generally not crucial in contexts where there is data abundance. 
	Although we do not focus on a specific application, we show that top-tuning methodology can be particularly effective when data availability is limited and fast training is relevant. This is typical of robotics devices and autonomous systems, especially when operating in unconstrained, quickly changing scenarios, where multiple training may need to be done on the fly\cite{maiettini2018speeding, DBLP:journals/corr/abs-1709-09882}. Our study shows that top-tuning is a promising alternative to fine-tuning for small-medium-sized datasets. Indeed, top-tuning provides comparable accuracy with respect to fine-tuning
	%(sometimes slightly worse, and sometimes slightly better)
	, with training times between one and two orders of magnitude smaller.\\
	
	More in detail, to assess the potential of the proposed methodology, in our study we focus on three different aspects: {\em (a) Target dataset:} to ensure the generality of our empirical observations, we consider 32 small to medium-sized target datasets, showing that the top-tuning approach is $\simeq85$ times faster on average with respect to the fine-tuning one. To provide more robust results we confirm our findings with two additional head classifiers: a naive shallow net and a ridge regressor, showing a smaller still significant speed-up. {\em (b) Pre-trained model:} to evaluate the influence of the specific pre-trained model,  we include seven different state-of-the-art pre-trained models. We consider complex architectures such as Xception and Vision Transformers, as well as smaller models for embedded devices such as MobileNetV2. Our findings are confirmed showing that despite different pre-trained models, the top-tuning approach is $\simeq70$ times faster on average with respect to the fine-tuning one. {\em (c) Pre-training dataset:} to assess dependency on the source dataset, we consider four distinct datasets for pre-training, taking into account different factors such as the amount of images and classes, and image size. We show the influence of the pre-training dataset on the downstream task, identifying the number of classes as a crucial factor in the pre-training dataset.\\
	
	The main contributions of this work are: (1) A systematic empirical study on training time efficiency involving 3500 training processes, comparing top-tuning and fine-tuning approaches on a large ensemble of small to medium size datasets. (2) An evaluation of a fast kernel classifier method based on the Nystr{\"o}m approximation when coupled with pre-trained features, showing its potential wide applicability to the image domain and its effectiveness in general-purpose computer vision scenarios. (3) An analysis of the effectiveness of pre-training quality, which prioritizes the number of classes to the number of samples.\\
	
	The rest of the paper is organized as follows:
	\autoref{ch:Related_works} presents an account of the relevant background while \autoref{ch:Methodology} reports details on our methodology. In \autoref{ch:Experiments design} we present the hyper-parameters configuration and we provide details on the datasets involved in the experiments. In \autoref{ch:Experiments}, we illustrate the results of our empirical analysis. Finally, \autoref{ch:Conclusions} is left to concluding remarks.

	\section{Related works}\label{ch:Related_works}
	%The investigation of transfer learning settings and how different parameters influence performances in terms of accuracy and training time is not itself a novelty. 
	In the last few years, different papers focused on the importance of convolutional features. For instance, in \cite{DBLP:journals/corr/Garcia-GasullaP17} the authors analyze how convolutional networks behave as feature extractors. An in-depth analysis is provided on the importance of low, middle, and high-level convolutional features. In \cite{sharif2014cnn} the effectiveness of CNN features is tested on a series of experiments conducted on recognition tasks. In \cite{9012507} different classification algorithms are tested on image features extracted via a convolutional neural network. Moreover, recent papers focus on the properties of specific neural network architectures on transfer-learning accuracy. In \cite{47104}, the authors evaluate if models that exhibit superior performance while being trained on ImageNet also display better performance in other visual tasks. They show that model accuracy on ImageNet strongly correlates with the performances of the same model on target tasks except for fine-grained image datasets, where pretraining on ImageNet provides minimal benefits.
	
	To assess the importance of the ImageNet pre-train, recent works focused on ImageNet factors that contribute to the superior descriptive abilities of pre-trained models' learned features. In \cite{DBLP:journals/corr/HuhAE16} the authors perform an empirical investigation on ImageNet properties, training a CNN on different subsets of ImageNet, varying the number of samples, classes, and granularity to assess their effect in terms of accuracy.
	
	Several works focus on the problem of speeding up the training process, evaluating potential influencing factors, and proposing faster training strategies.
	In \cite{visual_prompt_tuning} the authors propose a method to significantly reduce the training cost of a transformer-based model, by updating only a limited portion of the weights. In \cite{scaling_and_shifting} a parameter-efficient method is proposed based on scaling and shifting deep features extracted from a pre-trained model. In \cite{unified_parameter_efficient_transfer} the authors present a unified framework that establishes connections between the design of state-of-the-art parameter-efficient methods.
	In \cite{He_2015_CVPR} the authors focus on the impact of a CNN complexity on the training time, evaluating the accuracy with constrained time cost, considering the influence of different factors including depth, number of filters, and filter size.
	In \cite{Bigtransfer}, the authors investigate the paradigm of pre-training deep models on large supervised datasets in a transfer-learning framework for different vision tasks. They introduce a fine-tuning protocol that avoids expensive hyperparameters search, which is replaced by a custom heuristic procedure. Despite the recent attempts to reduce the costs associated with training procedures, the average number of parameters of the referenced models ranges from $10^5$ to $10^8$. Our approach, instead, usually involves $10^4$ training parameters, a number that is one order of magnitude smaller than the most efficient referenced fine-tuning protocol, and up to four orders of magnitude smaller than the others.
	% Moreover, in \cite{predicting_training_time} the authors propose an approach to predict the time needed by a deep model to be trained. They compute the time for different parts of a deep learning model. Then, by combining individual parts can then be combined the whole training time can be computed.
	
	Moreover, different strategies to reduce training and inference time for CNNs are proposed, including the combination of simplified convolutional filters\cite{mamalet} or the use of one-dimensional filters\cite{speedupcompression}. 
	Another relevant area of research includes the compression of deep neural networks\cite{cheng2017survey} to reduce their storage and computational inference cost. This research field focuses on different approaches such as parameters pruning\cite{248452} and quantization\cite{gholamisurvey}, aiming at decreasing the time the model needs to make a prediction. Indeed such techniques can be particularly useful to reduce computational costs at test time.
	
	Nonetheless, the aforementioned works focus on strategies to reduce training or inference time, with no specific comparison between different approaches nor evaluating the trade-off between accuracy and training resources. In this paper, instead, we perform an extensive training time and accuracy analysis, comparing different top-tuning and fine-tuning approaches on a large ensemble from small to medium size datasets, a relevant regime for transfer learning problems. We show that convolution features pre-trained on rich datasets such as ImageNet, provide general-purpose representation which can be transferred to a new task simply by training an external classifier. Such an approach shows a comparable accuracy with respect to fine-tuning, with a significant training process speed-up.
	\\
	The speed-up we achieve is also boosted by the choice of a fast kernel method as a head classifier. In the last few years, the fast kernel approach we are adopting was tested in different scenarios. In \cite{meanti2020kernel, musco2017recursive} theoretical guarantees for the Nystr{\"o}m approximation technique are provided and empirically examined. Its effectiveness in terms of speed is shown in classical numerical feature datasets, not with computer vision datasets. In \cite{maiettini2018speeding, ceola2022learn} the fast kernel is tested on specific robotic datasets for detection and pose-estimation tasks. In \cite{belkin2018understand, meanti2022efficient} Nystr{\"o}m approximation is tested on basic image datasets such as MNIST and CIFAR10 while in \cite{williams2021neural} it is used for 3D surface reconstruction. In our work we instead present an extensive analysis of more than thirty general-purpose image datasets, proving the effectiveness of the fast kernel approach on a wide set of image classification tasks, and showing its potential in the computer vision domain.
	
	\section {Theory} \label{ch:Methodology}
	We assume we are given a state-of-the-art deep learning model pre-trained on a source dataset (e.g. ImageNet), and a target dataset of input-output pairs $\{(x_i, y_i)\}_{i=1}^n$ to address a new image classification problem. The inputs  
	images, the outputs are discrete labels $y_i\in \{1, \dots, T\}$, for instance different semantic classes. In our study, we consider {two alternative transfer approaches, with the aim of highlighting the potential of our {\em top-tuning} methodology versus the common fine-tuning alternative:}
	\begin{enumerate}
		\item \textit{Fine-tuning}:  a typical deep learning architecture for image classification, {includes %a pre-trained part composed by 
			$C$ convolutional layers} followed by { L} fully connected layers, which we may formalize as:
		\begin{equation}\label{eq:cnn}
			\Phi_{FT}=\underbrace{ \Phi_{C+L} \circ \ldots \Phi_{C+1}}_{\text{Fully connected layers}}\ \circ \underbrace{\Phi_{C}\circ \ldots  \circ \Phi_{1}(x)}_{\text{Convolutional layers}}.
		\end{equation}
		In the basic idea of transfer learning by {\em fine-tuning}, we start from a pre-trained model $\Phi_{PT}$ and update the structure of the fully connected layers to address the new classification task. 
		Then we fine-tune the model on the target dataset. 
		Notice that this procedure may involve both the parameters of the fully connected and of the convolutional layers in an end-to-end fashion minimizing an empirical error via back-propagation. {It can be more or less deep, depending on the portion of the architecture we want to fine-tune}. The fine-tuning pipeline is shown in \autoref{fig:Pipeline}(Left). 
		
		\item \textit{Top-tuning}: in this case, we start from a pre-trained model as in Equation \ref{eq:cnn}; we fix the convolutional layers $\Phi_{C}, \ldots,\Phi_{1}$. We then compute the pre-trained convolutional features on the target dataset and use them to train a fast kernel classifier. 
		The overall  top-tuning pipeline is shown in \autoref{fig:Pipeline}(Right), and
		may be formalized as follows:
		\begin{equation}\label{eq:kcnn}
			\Phi_{TT}= \underbrace{ \Psi}_{\text{Kernel feature map}}\circ \underbrace{\Phi_{C}\circ \ldots  \circ \Phi_{1}(x)}_{\text{Convolutional layers}}.
		\end{equation}
		$\Psi$ is a typically infinite-dimensional feature map corresponding to a kernel $k(z,z')$, where $z$ and $z'$ are feature vectors composed by the pre-trained convolutional features computed as 
		
		$$z=\Phi_{C}\circ \ldots  \circ \Phi_{1}(x).$$
		
		Such features are assumed to be fixed and are not trained/tuned. Therefore, the only free parameters $W$ are computed by a ridge regression minimizing
		\begin{equation}\label{eq:KRS}
			\hat{f}(z) = \min_W\sum_{i=1}^n\|W z_i-\mathbf y _i\|^2+\lambda \|W\|_F^2
		\end{equation}
		where $\|\cdot\|_F$ is the Frobenious norm, $\lambda$ is a regularization parameter controlling the complexity of the solution. Using the representer theorem \cite{scholkopf2001generalized} \autoref{eq:KRS} can be re-written as
		\begin{equation}
			f(z) = W\Phi(z)= \sum_{i=1}^n k(z,z_i) \boldsymbol \alpha_i, \qquad \boldsymbol \alpha_i\in \mathbb R^T,
		\end{equation}
		where the coefficients $\alpha_i$ satisfy the linear system
		\begin{equation}\label{KRR}
			(\mathbf K +\lambda I)\boldsymbol \alpha = \mathbf Y, 
		\end{equation}
		with $\boldsymbol \alpha = (\boldsymbol \alpha_1, \dots, \boldsymbol \alpha_n), \mathbf Y = (\mathbf y_1, \dots, \mathbf y_n) \in \mathbb R^{nT},$ and $\mathbf K\in \mathbb R ^{nT}$ with 
		$\mathbf K _{ij}=k(x_i,x_j)$. The solution can be determined as
		\begin{equation}\label{eq:alphas_kernel}
			\boldsymbol \alpha = (\mathbf{K} + \lambda \mathbf{I} )^{-1} \mathbf{y}.
		\end{equation}
		\autoref{eq:alphas_kernel} requires inverting an $n \times n$ matrix, with $n$ training samples; this operation  cost $\mathcal{O}(n^3)$ in time and $\mathcal{O}(n^2)$ in space. To make the kernel ridge classifier significantly faster and more efficient, we rely on Nystr{\"o}m approximation\cite{nystrom} that solves the same problem in a smaller space. Therefore we consider a reduced set of $m$ \textit{inducing points} taken among the $n$ of the training set. By taking $m=\mathcal{O}(\sqrt{n})$ we have strong theoretical guarantees and empirical evidence that the error does not increase and leads to computational cost reduction. The time cost is reduced to $\mathcal{O}(n\ log(n)\ \sqrt{n})$. The space needed also decreases to $\mathcal{O}(n)$. More details about this procedure can be found in\cite{rudi2018falkon, meanti2020kernel}. 
	\end{enumerate}
	
	\section{Material and methods}\label{ch:Experiments design}
	In this section, we present technical details about the conducted experiments, to allow the work to be reproduced by an independent researcher. First, we consider the hyper-parameters tuning, both for the fine-tuning and top-tuning procedures. Then we report details about the datasets used in the empirical analysis.
	
	\subsection{Hyper-parameters tuning}\label{ch:HPT}
	A model can be tuned in several ways involving numerous hyper-parameters. In our analysis, we do not focus on an exhaustive hyper-parameters exploration. Instead, we consider a set of plausible configurations, both for the fine-tuning and the top-tuning pipelines, according to guidelines in previous works. It is worth noticing that, for each pipeline, the overall training time is computed as the summation of every considered configuration training time.\\
	\\
	\noindent\textbf{Fine-tuning hyper-parameters:}\\
	We perform a five-fold cross-validation analysis involving the following hyper-parameters:
	\begin{itemize}
		\item[-]Training steps: inspired by previous studies on a similar context\cite{47104} we limited this quantity to 20.000 training steps, coupled with an early stopping criterion.
		\item[-]Early stopping: we monitor the validation loss with a patience parameter equal to $10$.
		\item[-]Weights update: we only consider fine-tuning the whole convolutional part, making the model more adaptable to the downstream task with respect to the top-tuning approach. That is to avoid a combinatorial explosion in the number of configurations, that would result in excessive and unfair training time for the fine-tuning pipeline.
		\item[-]Batch size: taking into account previous studies about batch size\cite{he2019control, masters2018revisiting, bengio2012practical} we compute it as: $b=\lfloor 2^{2 \cdot log_{10}(n)-1} \rfloor$ where $n$ is the number of points in the dataset.
		\item[-]Optimizer: we use default Stochastic Gradient Descent (SGD). As suggested by\cite{Goodfellow-et-al-2016}, we use two different learning rates: $l = \{0.1, 0.01\}$
	\end{itemize}
	Overall, we run two training configurations for every fold, one for each considered learning rate value as all the remaining hyper-parameters are fixed. To evaluate the accuracy, we consider the best-performing configuration. Training time is the sum of the training times requested for both configurations.\\ \\
	
	\noindent\textbf{Top-tuning hyper-parameters}:\\
	Also for the top-tuning approach, we perform a five-fold cross-validation analysis involving the following hyper-parameters:
	\begin{itemize}
		\item[-] Kernel: the approximated kernel ridge regression is based on a Gaussian kernel $k(z,z')= e^{-\frac{\|z-z'\|^2}{2\gamma^2}}$, with $\gamma$ kernel width. We use two values of $\gamma = \{10^2, 10^3\}$.
		\item[-] Regularization: we consider two values for the regularization term: $\lambda = \{10^{-5}, 10^{-6}\}$
	\end{itemize}
	Overall, we run four training configurations for every fold, as two values for both kernel width and regularization are tested. To evaluate the accuracy, we consider the best-performing configuration. Training time is the sum of the training times requested for the four configurations.\\
	The described hyperparameter tuning procedure is based on a coarse grid search. This allows us to perform a fair comparison with fine-tuning approaches, where the number of hyperparameters is much larger.  However, as an alternative, it is possible to exploit automatic hyperparameter optimization procedures for fast-kernel approaches, as the one outlined in \cite{meanti2022}, developed for Nystr{\"o}m approximation methods.
	
	% We perform experiments automatic hyper-parameter optimization generally brings lower accuracy with respect to the grid search procedure described above. Therefore, in the experiments section, we present the results obtained through grid search strategy.
	
	\subsection{Datasets details}
	\begin{table*}[!b]
		\centering
		\setlength{\tabcolsep}{1pt}
		\begin{tabular}{cccc}
			\hline
			Dataset name & \hspace*{2mm}\#images (Tr/Te) \hspace*{2mm} & Img. size mean & \#classes\\
			\hline
			AFHQ (AF)\cite{Choi_2020_CVPR} & 13.167/1.463 & $512 \times 512$ & 3\\
			Beans (BE)\cite{beansdata} & 1.167/128 & $500 \times 500$& 3\\
			Best artworks (BA)\cite{best_artworks}& 7.896/878 & $980 \times 921$& 50\\
			Boat types (BT)\cite{boat_types} & 1.315/147 & $905 \times 1234$ & 9\\
			Caltech-101 (C101)\cite{Fei-Fei2007} & 3.060/6.084 & $251 \times 282$ & 102\\
			Cassava (CSV)\cite{mwebaze2019icassava} & 7.545/1.885 & $573 \times 611$ & 5\\
			Cats vs Dogs (CVSD) \cite{asirra2007} & 20.935/2.327 & $365 \times 410$& 2\\
			Chest xray (CXRAY) \cite{kermany2018identifying} & 4.708/524 & $968 \times 1321$& 2\\
			CIFAR10 (CIF10) \cite{Krizhevsky09learningmultiple} & 50.000/10.000 & $32\times32$ & 10\\
			CIFAR100 (CIF100) \cite{Krizhevsky09learningmultiple} & 50.000/10.000 & $32\times32$ & 100\\
			Citrus leaves (CLV) \cite{rauf2019citrus} & 534/60 & $256 \times 256$& 4\\
			Colorectal hist (COL) \cite{kather_zollner_bianconi_melchers_schad_gaiser_marx_weis_2016} & 4.500/500 & $150 \times 150$& 8\\
			Deep weeds (DW) \cite{DeepWeeds2019} & 15.758/1.751 & $256 \times 256$ & 9\\
			DTD (DTD)\cite{cimpoi14describing} & 3.760/1.880& $ 453\times 500$ & 47\\
			EuroSAT (ES) \cite{helber2019eurosat} & 24.300/2.700& $64 \times 64$ & 10\\
			FGVC Aircraft (AIR)\cite{maji13fine-grained} & 6.667/3.333 & $353 \times 1056 $ & 100\\
			Footb vs Rugby (FVSR) \cite{football_rugby} & 2.203/245 & $618 \times 788$& 2\\
			Gemstones (GEM) \cite{gemstones_images} & 2.571/286 & $330 \times 335$& 87\\
			Hors or Hum (HVSH) \cite{horses_or_humans} & 1.027/256 & $300 \times 300$ & 2\\
			iCubWorld subset (ICUB)\cite{DBLP:journals/corr/abs-1709-09882} & 86.400/9.600 & $256 \times 256 $ & 10\\
			Indian Food (IF) \cite{indian_food} & 3.600/400 & $550 \times 610$ & 80\\
			Make No Make(MVSN)\cite{makeup} & 1.355/151 & $211 \times 246$& 2\\
			Malaria (MAL) \cite{rajaraman2018pre} & 24.802/2.756 & $133 \times 132$& 2\\
			Meat quality (MQA) \cite{8946388} & 1.706/190 & $720 \times 1280$& 2\\
			Oxford Flowers (OF) \cite{Nilsback08} & 2.040/6.149 & $538 \times 624$& 102\\
			Oxford-IIIT Pets (OP) \cite{parkhi12a} & 3.680/3.669 & $383 \times 431$ & 37\\
			Plankton (PL) \cite{Pastore856815} & 4.500/500 & $106 \times 120$ & 10\\
			Sars Covid (SCOV) \cite{DVN/SZDUQX_2020} & 2.232/249 & $260 \times 350$& 2\\
			Stanford Cars (SC) \cite{KrauseStarkDengFei-Fei_3DRR2013} & 8.144/8.041 & $308 \times 573 $& 196\\
			Stanford Dogs (SD) \cite{Khosla2012NovelDF} & 12.000/8.580 & $386 \times 443$& 120\\
			Tensorflow Flowers(TFF) \cite{tfflowers} & 3.303/367 & $272 \times 365$& 5\\
			Weather (MW) \cite{weather} & 1.012/113 & $335 \times 506$& 4\\
			\hline
		\end{tabular}
		\caption{The datasets adopted in our analysis. For every dataset, we provide the number of images (train and test split, respectively), the mean image size of the dataset, and the number of classes.}
		\label{table:dataset_infos}
	\end{table*}
	We include 32 small to medium-sized datasets in our experiments, from a wide range of contexts and scenarios (see \autoref{table:dataset_infos}). Our collection includes popular datasets in the computer vision community, like CIFAR10/100 and Caltech101, as well as more challenging datasets where the amount of data is limited with respect to the number of classes and complexity of the task (e.g., Stanford Cars and FGVC aircraft).
	Finally, we include datasets with a significantly limited amount of images, to consider practical cases where transfer learning procedures may be fundamental (e.g., Beans and Citrus leaves). Indeed, we did not include large datasets in our analysis since the transfer learning technique is generally not crucial in contexts where there is data abundance.
	From \autoref{table:dataset_infos} we can notice that both numbers of images and classes can vary deeply from one task to another. This can influence the task's hardness. On average each dataset has $11746.46$
	% \pm 18237.7$
	images and $35.21$
	% \pm 48.32$
	classes. The average number of images per class is $1780.24$
	% \pm 3077.48$. It is worth noticing the very high standard deviation. 
	The dataset with the most images per class is the Malaria dataset with 12401 images per class. The dataset with few images per class is Oxford Flowers with 20 images per class.

	\section{Results}\label{ch:Experiments}
	We now describe the details of the empirical analysis. Firstly, we conduct a comparison between fine-tuning and top-tuning, focusing on accuracy and training time. Then, to ensure robustness in our findings, we substitute the fast-kernel head classifier with a vanilla fully connected network or a ridge regressor. Later, to assess the impact of the pre-trained architecture on the results, we explore seven different pre-trained neural networks as potential replacements for the backbone used in both fine-tuning and top-tuning approaches. Lastly, we evaluate the significance of the pre-trained source data by considering four different datasets for pre-training the convolutional backbone used in both fine-tuning and top-tuning. All the experiments have been carried out on a single Quadro RTX 6000 GPU, 24Gb VRAM.
	
	\subsection{Top-tuning is highly faster with limited accuracy drop}\label{sec:Analysis on different datasets}
	To compare fine-tuning and top-tuning approaches, we first fix the neural network model and the pre-training dataset, considering an ImageNet pre-trained DenseNet201\cite{huang2018densely} architecture. This model represents a good compromise between predictive power and size, in terms of the number of parameters.
	We consider the 6 different configurations defined in \autoref{ch:HPT} averaging the results of a 5-folded procedure over the 32 datasets, resulting in $6\cdot5\cdot32=960$ distinct training processes.
	
	\begin{figure*}[!t]
		\centering
		{\includegraphics[width=0.9\textwidth]{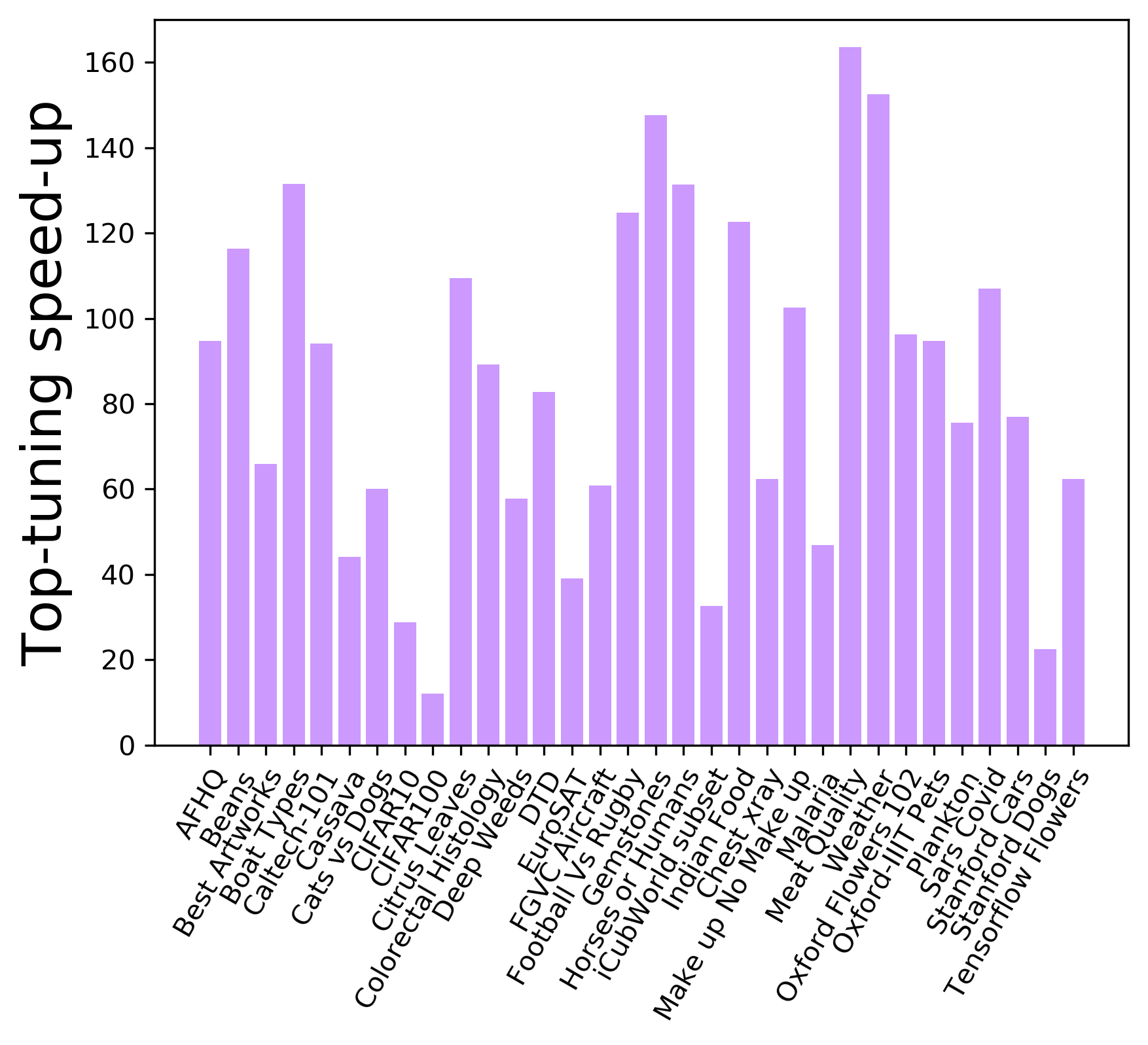}}
		\caption{Overall speed-up on different target datasets obtained by the top-tuning model with respect to the fine-tuning one. Each column represents a dataset. The corresponding height represents the speed-up factor of the top-tuning model with respect to the fine-tuning one(e.g. on the AFHQ dataset the top-tuning training was $\sim95$ times faster). All the experiments have been carried out on a single Quadro RTX 6000 GPU, 24Gb VRAM.}
		\label{fig:Datasets_times}
	\end{figure*}
	
	In \autoref{fig:Datasets_times} we show the overall results in terms of training time speed-up. Each column refers to a different dataset, reporting the speed-up $SpUp$ obtained by the top-tuning with respect to fine-tuning. 
	%For instance, on the AFHQ dataset, the top-tuning training was $\sim 95$ times faster. 
	The top-tuning model is always highly faster to train than the fine-tuning one. Indeed, given the fine-tuning training time $t_{\text{fine-tuning}}$ and the top-tuning training time $t_{\text{top-tuning}}$,
	
	$$SpUp=\dfrac{t_{\text{fine-tuning}}}{t_{\text{top-tuning}}} $$ %\in }
	
	is in the range $[10,165]$, with the top-tuning model being between 10 and 165 times faster to train. The average speed-up is $84.64\pm38.97$ across the datasets.
	%On average, on a single training process, the time needed by the fine-tuning method required $\sim 48$ minutes, while the top-tuning approach needed $\sim1$ minute. 
	On larger datasets, e.g. CIFAR100, the training time was reduced from $\sim2$ hours to $\sim10$ minutes; computed as the training time sum of the two fine-tuning and four top-tuning configurations, respectively.
	
	We can relate the faster training time to: (i) number of parameters: the top-tuning model has a number of parameters on average two orders of magnitude lower than the fine-tuning ones. Indeed, a fully tunable convolutional model like DenseNet contains more than 20 million learnable parameters. Instead, the top-tuning model is made by a few dozen thousand parameters (ii) impact of backpropagation: every layer needs to wait for the subsequent layer computation.\\
	
\begin{figure*}[!t]
		\centering
		{\includegraphics[width=0.8\textwidth]{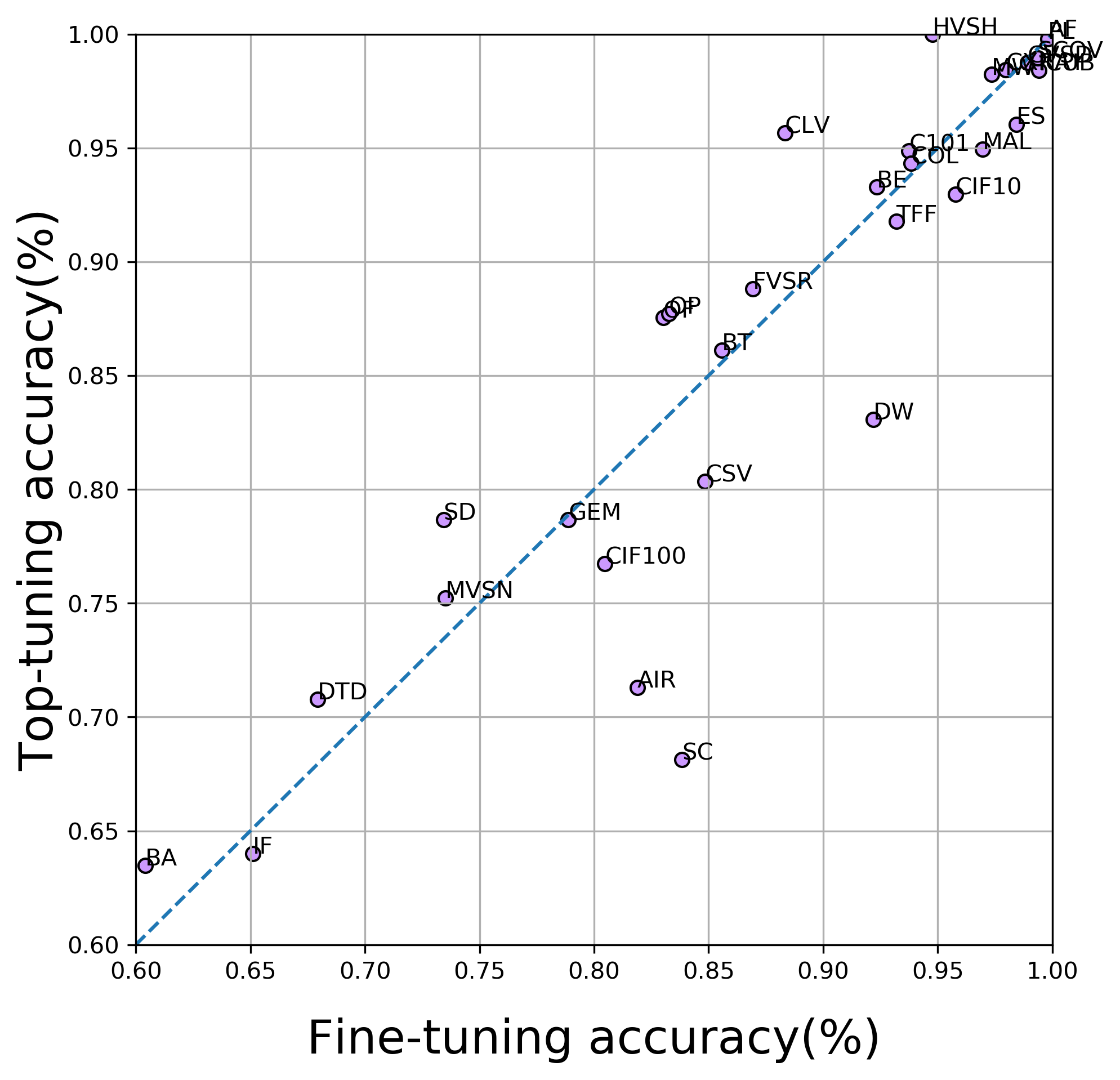}}
		\caption{Overall accuracies on different target datasets. Each point represents a dataset. The accuracy obtained by the best fine-tuning model is shown on the x-axis. The best top-tuning model accuracy is shown on the y-axis. When a point is lying below the diagonal, the fine-tuning model is performing better w.r.t the top-tuning one, and vice-versa.}
		\label{fig:Datasets_accs}
	\end{figure*}

	To show that training time speed-up does not affect accuracy,  in \autoref{fig:Datasets_accs} we report the overall accuracy results. Each point represents a different dataset. Its position is given by the accuracy obtained by the best fine-tuning configuration on the x-axis, and by the best top-tuning configuration on the y-axis. The diagonal is marked for readability purposes. Intuitively, when a point is lying below the diagonal, the fine-tuning model is performing better w.r.t the top-tuning one, and vice-versa. 
	
	\noindent Most of the datasets lie around the diagonal, showing similar accuracy between fine-tuning and top-tuning methods. Indeed, 
	$$\Delta Acc = Acc_{\mathrm{top-tuning}} - Acc_{\mathrm{fine-tuning}} $$ 
	
	\noindent is in the range $[ -2.5\%,+2.5\%]$ in $60\%$ of our experiments.
	%For these datasets, the benefits coming from the fine-tuning procedure are either absent or typically marginal. Indeed, on half of the considered datasets, the top-tuning procedure is performing better than the fine-tuning one.
	Only on a few datasets, e.g. FGVC aircraft and Stanford Cars(SC), fine-tuning provides a  benefit.
	This behavior could be related to two factors. (i) Representation in ImageNet: as pointed out by \cite{47104},  in ImageNet cars and planes are represented at a coarse-grained level. For instance, ImageNet contains only two plane classes; (ii) Dataset hardness in terms of the number of classes, granularity, and the number of images per class.

	%to proceed with its update.
	\begin{table*}[!t]\setlength{\tabcolsep}{2pt}
		\centering
		\begin{tabular}{ccc}
			\hline
			Dataset&\hspace*{1mm}$\Delta$Acc$\uparrow$\hspace*{2mm}&SpUp$\uparrow$\\
			\hline
			AFHQ  & $+0.10\%$ & $ 94.70 \times$ \\
			Beans  & $+0.90\% $ & $ 116.3 \times$ \\
			Best artworks  & $+3.10\% $ & $ 65.90 \times$ \\
			Boat types  & $+0.50\% $ & $ 131.5 \times$ \\
			Caltech-101  & $+1.10\% $ & $ 94.00 \times$ \\
			Cassava  & $-4.50\% $ & $ 44.00 \times$ \\
			Cats vs Dogs  & $-0.20\% $ & $ 60.00 \times$ \\
			Chest xray  & $+0.50\% $ & $ 62.40 \times$ \\
			CIFAR10  & $-2.80\% $ & $ 28.90 \times$ \\
			CIFAR100  & $-3.70\% $ & $ 12.10 \times$ \\
			Citrus leaves  & $+7.30\% $ & $ 109.5 \times$ \\
			Colorect. histology  & $+0.50\% $ & $ 89.20 \times$ \\
			Deep weeds  & $-9.10\% $ & $ 57.70 \times$ \\
			DTD  & $+2.90\% $ & $ 82.80 \times$ \\
			EuroSAT  & $-2.40\% $ & $ 39.10 \times$ \\
			FGVC Aircraft  & $-10.6\% $ & $ 60.90 \times$ \\
			\hline
		\end{tabular}
		\begin{tabular}{ccc}
			\hline
			Dataset&\hspace*{1mm}$\Delta$Acc$\uparrow$\hspace*{2mm}&SpUp$\uparrow$\\
			\hline
			Football vs Rugby  & $+1.90\% $ & $ 124.8 \times$ \\
			Gemstones & $-0.20\% $ & $ 147.6 \times$ \\
			Horses or Humans  & $+5.20\% $ & $ 131.3 \times$ \\
			iCub World subset & $-1.00\% $ & $ 32.60 \times$ \\
			Indian Food  & $-1.10\% $ & $ 122.6 \times$ \\
			Make up  & $+1.70\% $ & $ 102.5 \times$ \\
			Malaria  & $-2.00\% $ & $ 46.80 \times$ \\
			Meat Quality  & $+0.00\% $ & $ 163.6 \times$ \\
			Oxford Flowers102  & $+4.50\% $ & $ 96.30 \times$ \\
			Oxford-IIIT Pets  & $+4.50\% $ & $ 94.60 \times$ \\
			Plankton  & $+0.00\% $ & $ 75.50 \times$ \\
			Sars Covid  & $-0.40\% $ & $ 107.0 \times$ \\
			Stanford Cars  & $-15.7\% $ & $ 76.90 \times$ \\
			Stanford Dogs  & $+5.20\% $ & $ 22.50 \times$ \\
			Tensorflow flowers  & $-1.40\% $ & $ 62.40 \times$ \\
			Weather  & $+0.90\% $ & $ 152.4 \times$ \\
			\hline
		\end{tabular}
		\caption{Quantitative results about the analysis on different datasets. The second column reports the $\Delta Acc = Acc_{\mathrm{top-tuning}} - Acc_{\mathrm{fine-tuning}}$. The third one refers to the corresponding speed-up obtained when using the top-tuning procedure(e.g. on the AFHQ dataset the top-tuning training was $94.70$ times faster).}
		\label{table:Datasets_accs_times}
	\end{table*}
	\autoref{table:Datasets_accs_times} summarizes the quantitative results obtained with our experiments both in terms of accuracy and training time speed-up. For every dataset, we report the $\Delta Acc$ 
	% = Acc_{\mathrm{top-tuning}} - Acc_{\mathrm{fine-tuning}}$ 
	and the corresponding speed-up $SpUp$.\\
	Lastly, although in this work we do not focus on inference time, we report for completeness that the two pipelines needed similar time for a prediction.
	
	\subsection{Analysis with different head classifiers}
	To test the generality of our approach we replace the fast kernel classifier with two head classifiers: a shallow net and a ridge regression classifier. 
	The training procedure is similar to the one presented in \autoref{ch:Methodology}. For the shallow net, we consider just two configurations with default Stochastic Gradient Descent (SGD) and two different learning rates: $l = \{0.1, 0.01\}$. The training time reported in the results is the sum of the computational time required for the two training instances. Notice that this architecture is identical to the fine-tuning one. The main difference lies in which part of the model is tuned. The shallow nets update only the parameters of the last three fully connected layers while the fine-tuning one update all the weights of the model.
	
	For the Ridge regressor, we use three different configurations corresponding to three different values of the regularization term $\alpha=\{10^1, 10^{-1}, 10^{-3}\}$. Here again, the training time reported in the results is the sum of the computational time required for the three training instances.
	
\begin{figure*}[!t]
		\centering
		{\includegraphics[width=0.485\textwidth]{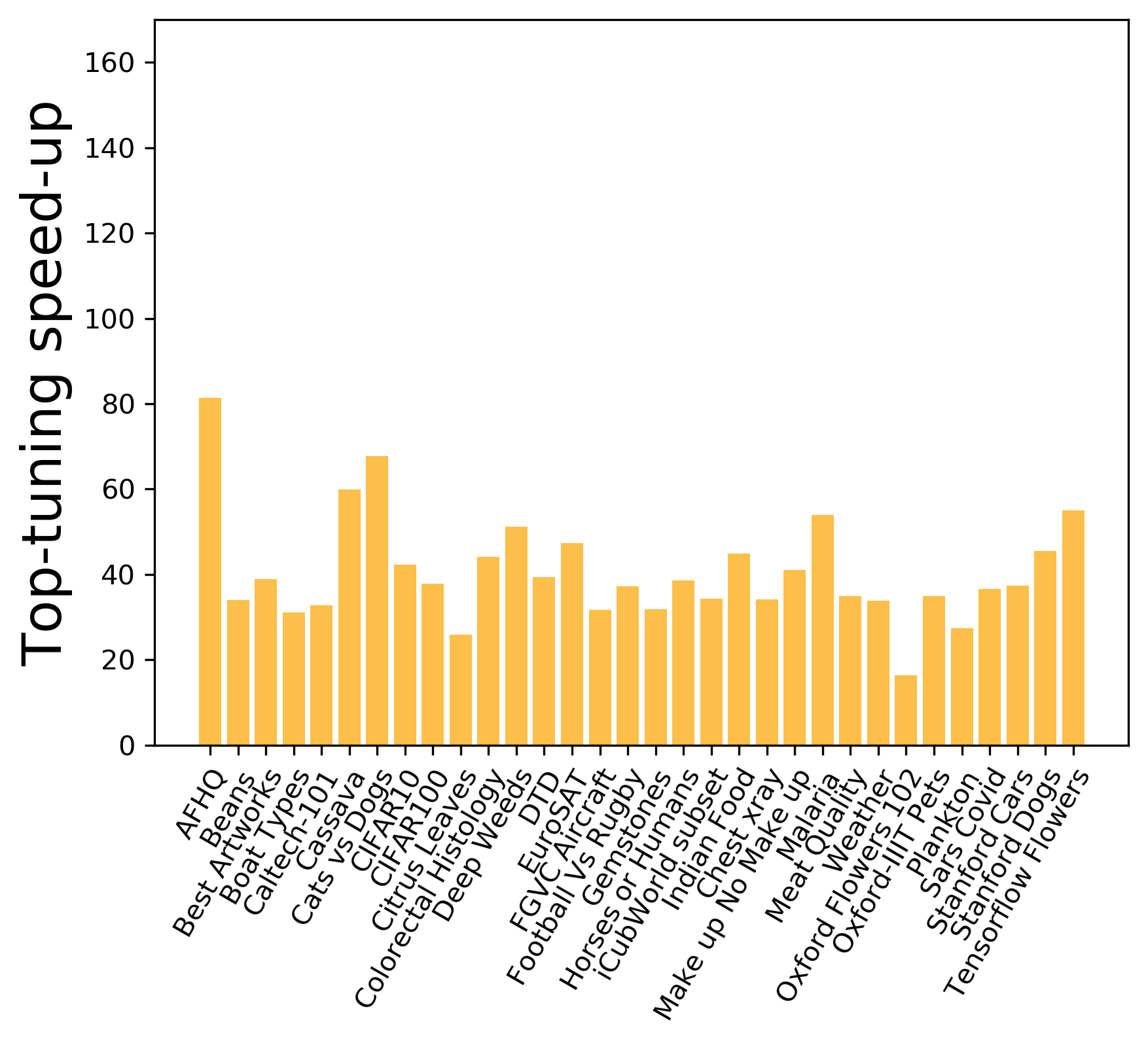}}
		\hspace{1mm}
		{\includegraphics[width=0.485\textwidth]{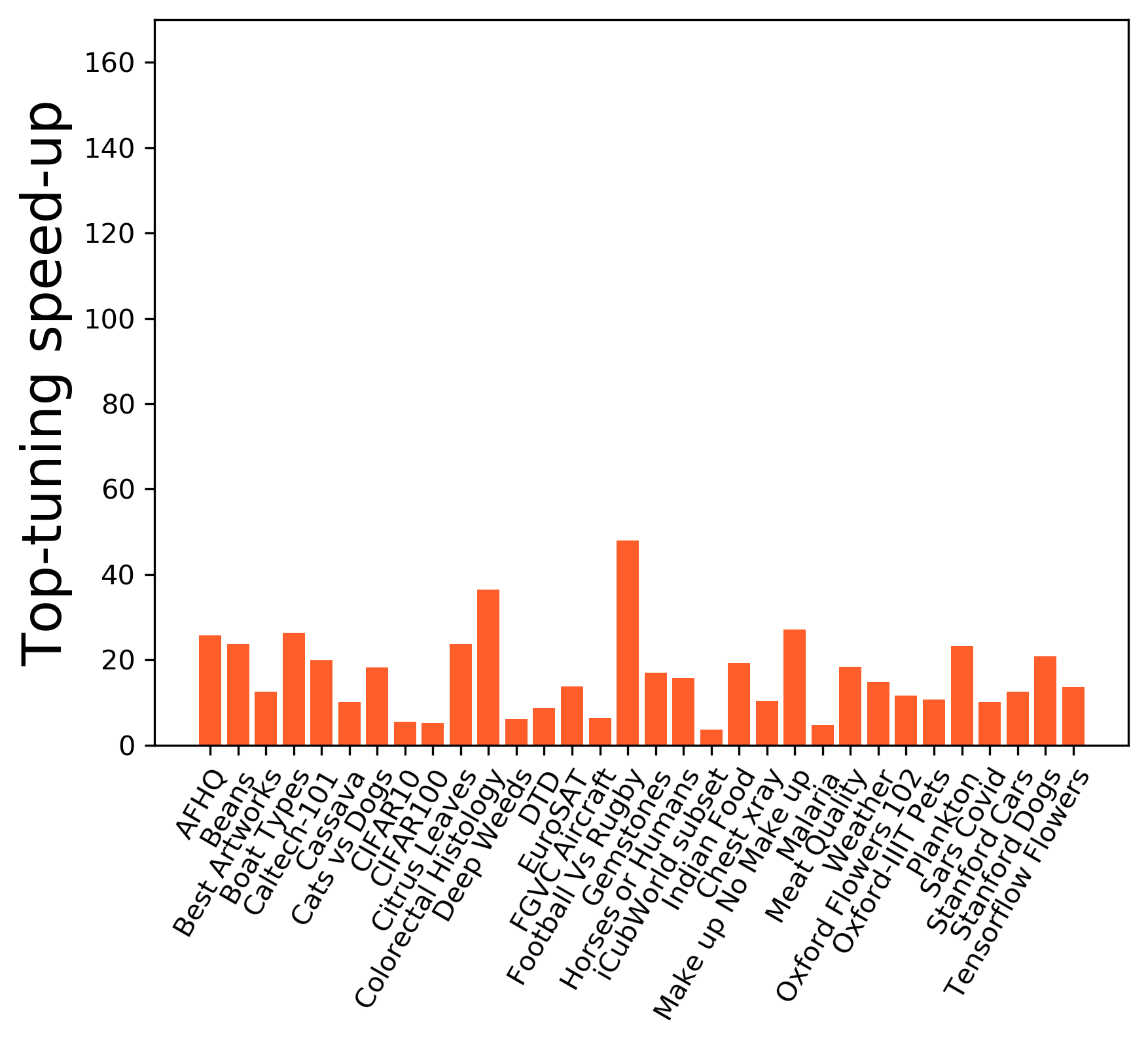}}
		\caption{Overall speed-up results on different target datasets with shallow net and ridge as external classifiers w.r.t the fine-tuning model. (Left) Speed-up obtained over different target datasets by using a shallow net as an external classifier. (Right) Speed-up obtained over different target datasets by using a ridge regressor as an external classifier.}
		\label{fig:Datasets_times_times}
	\end{figure*}

	In \autoref{fig:Datasets_times_times} we compare, in terms of training time, the fine-tuning model with both the shallow net(left) and the ridge regressor(right) as an external classifier. With the fast kernel classifier as an external classifier, the average speed-up is $84.64\pm38.97$. Instead, by using a shallow net as an external classifier we obtain an average speedup of $40.74 \pm 12.43$. The speed-up is instead $16.37 \pm 9.57$ if we use the ridge regressor as an external classifier. 
	
	These results show that, regardless of the choice of external classifier, top-tuning confirms a training time speed-up of at least one order of magnitude with respect to the fine-tuning pipeline. Among them, the fast-kernel model reports a significant boost with respect to to its competitors.
	
\begin{figure*}[!t]
		\centering
		{\includegraphics[width=0.485\textwidth]{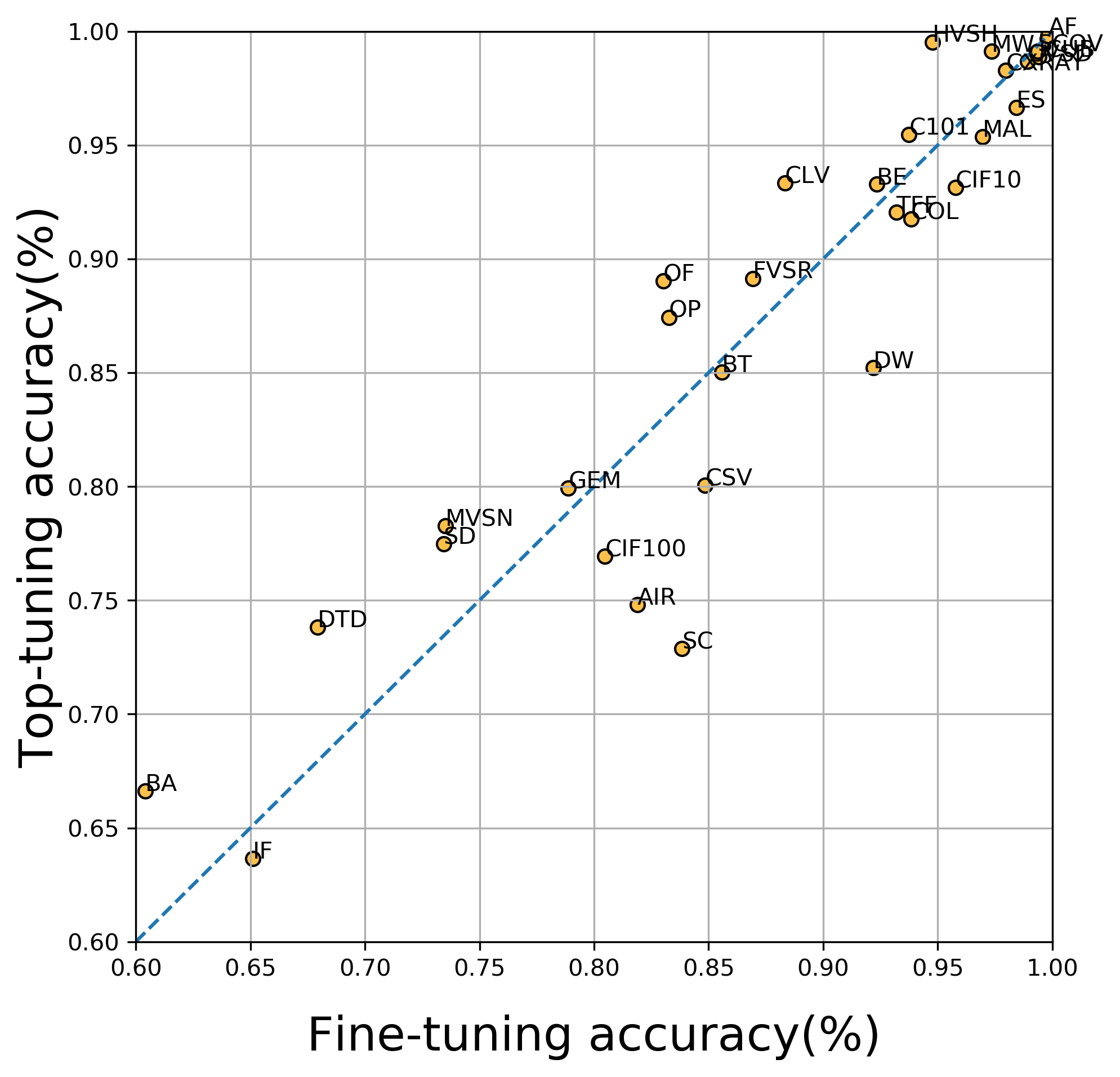}}
		\hspace{1mm}
		{\includegraphics[width=0.485\textwidth]{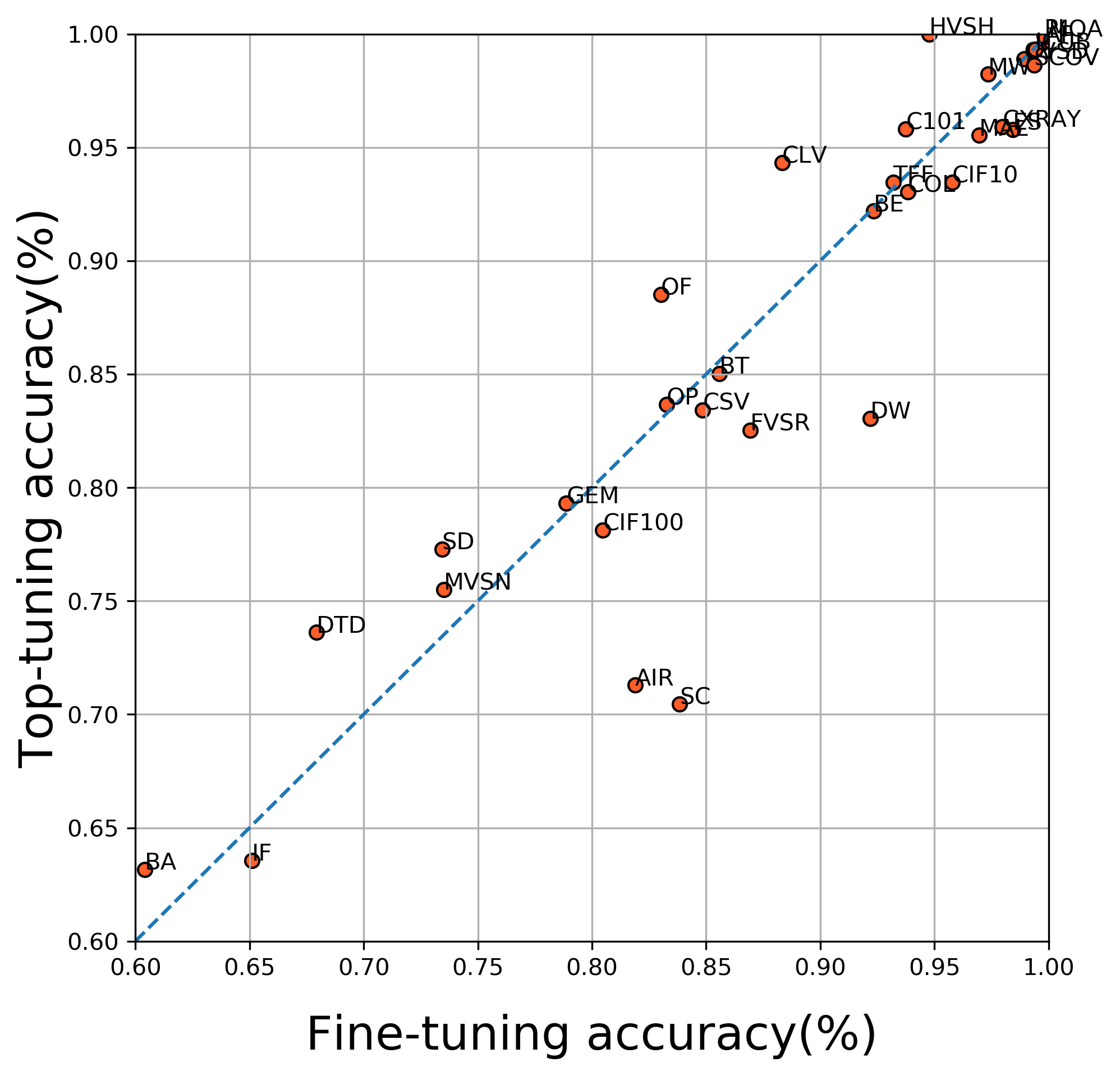}}
		\caption{Overall accuracy results on different target datasets with shallow net and ridge as external classifiers w.r.t the fine-tuning model. (Left) Accuracies obtained over different target datasets by using a shallow net as an external classifier. (Right) Accuracies obtained over different target datasets by using a ridge regressor as an external classifier.}
		\label{fig:Datasets_accs_accs}
	\end{figure*}

	In \autoref{fig:Datasets_accs_accs} we compare, in terms of accuracy, the fine-tuning model with both the shallow net(left) and the ridge regressor(right) as external classifiers. The obtained results are similar to the ones shown in Figure 2. Most of the datasets lie around the diagonal for both shallow net and ridge as an external classifier, showing similar accuracy between fine-tuning and top-tuning methods, confirming the findings of \autoref{sec:Analysis on different datasets}.
	%
	
	% \subsection{Results with different pre-trained neural networks.}\label{para: Analysis on different models}
	\subsection{Impact of pre-trained model}\label{para: Analysis on different models}
\begin{figure*}[!t]
		\centering
		{\includegraphics[width=0.47\textwidth]{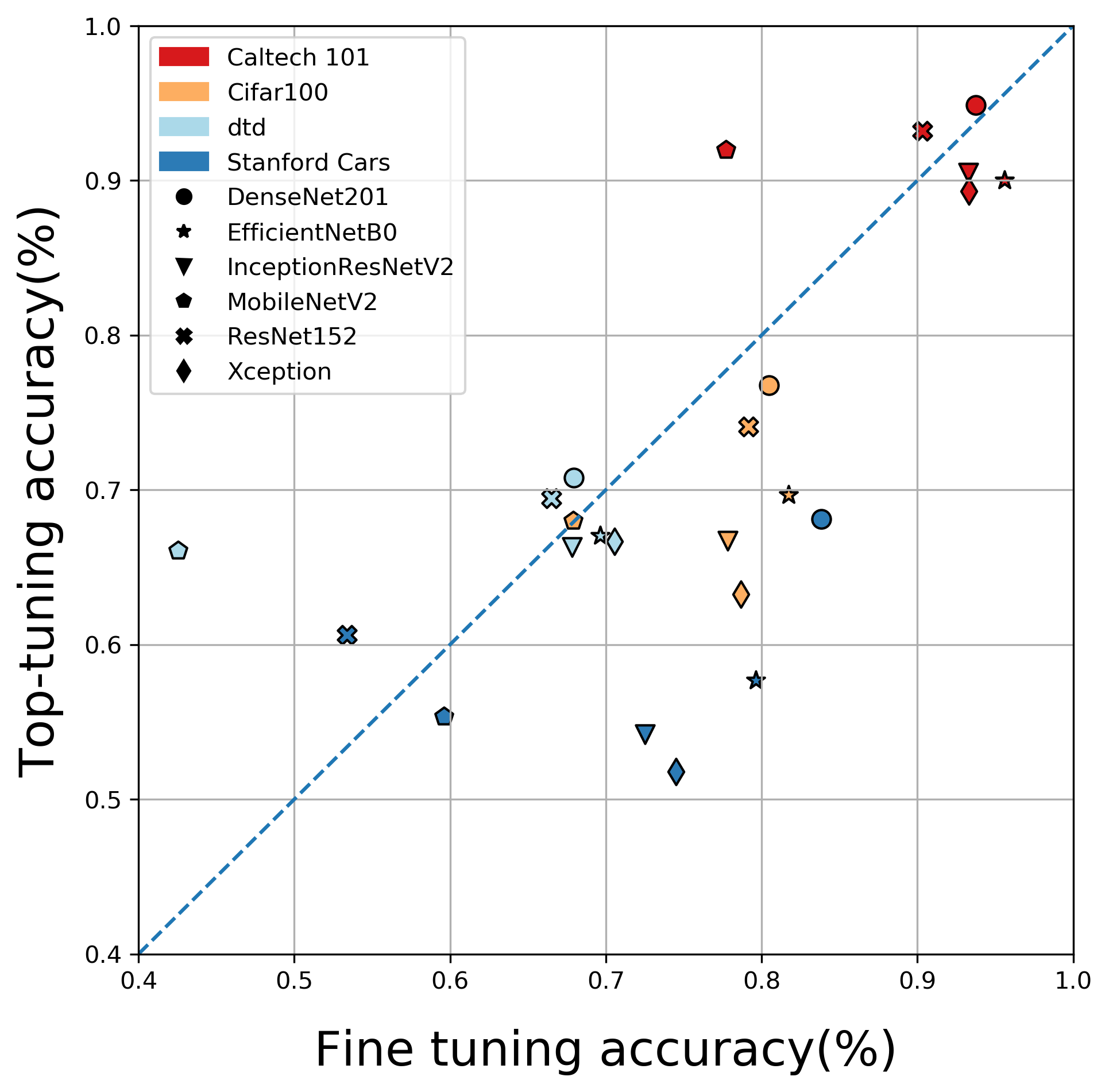}}
		\hspace{3mm}
		{\includegraphics[width=0.47\textwidth]{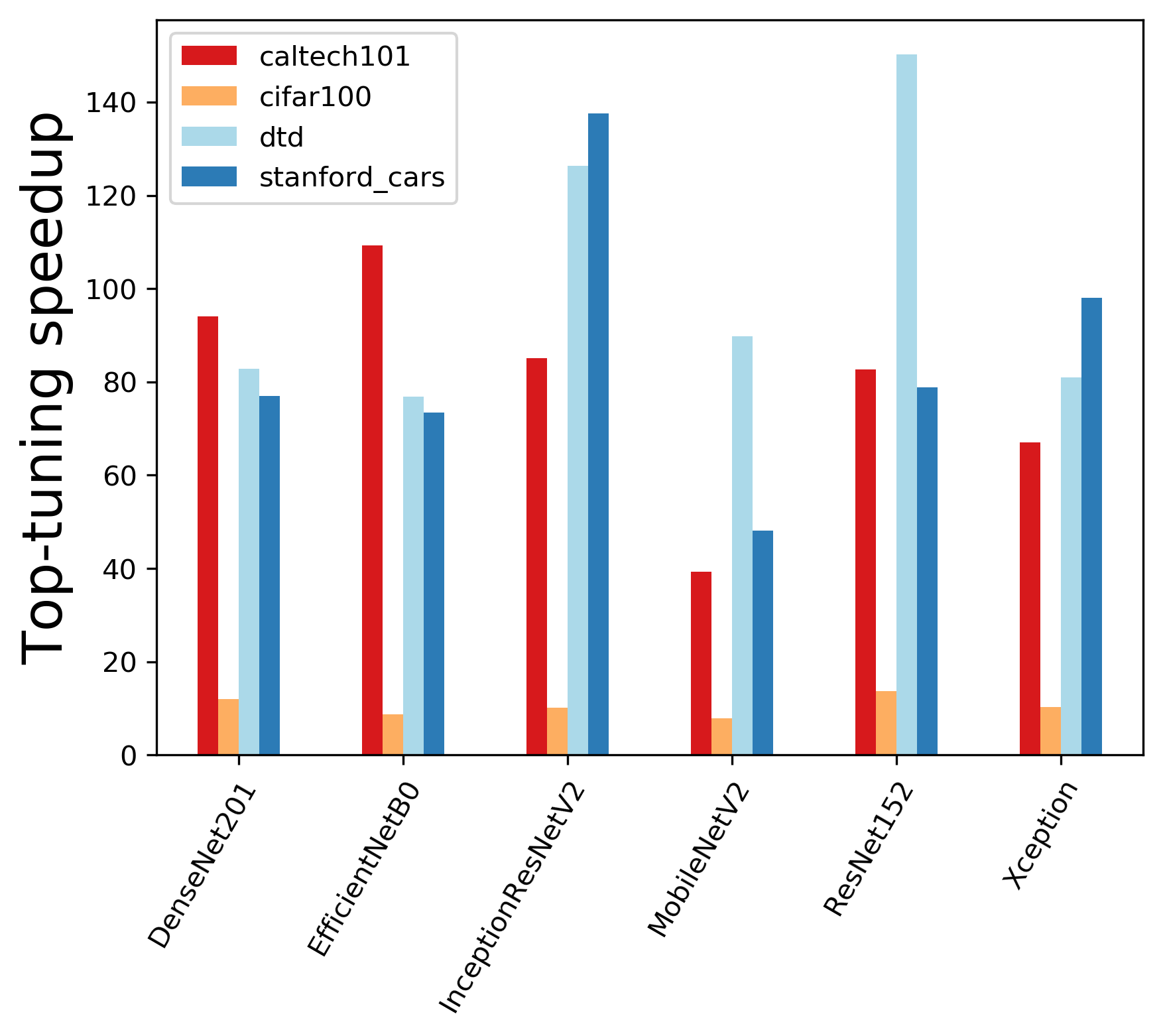}}
		\caption{Overall results on different pre-trained models. (Left) Accuracies over different target datasets with different pre-trained models. (Right) Speed-up obtained by the top-tuning model w.r.t the fine-tuning one for each dataset and for each pre-trained model.\vspace*{3mm}}
		\label{fig:Models_accs_times}
	\end{figure*}

	To weight the dependency of the results on the pre-trained architecture we extend the results obtained with DenseNet201 on five state-of-the-art pre-trained models: (i) EfficientNetB0 \cite{DBLP:journals/corr/abs-1905-11946}; (ii) InceptionResNetV2 \cite{DBLP:journals/corr/SzegedyIV16}; (iii) MobileNetV2\cite{DBLP:journals/corr/HowardZCKWWAA17}; (iv) ResNet152\cite{DBLP:journals/corr/HeZRS15}; (v) Xception\cite{DBLP:journals/corr/Chollet16a}.\\
	We test these models on four different datasets where: fine-tuning has only marginal benefits (Caltech-101, CIFAR100), top-tuning approach provides better results with respect to fine-tuning (DTD), and fine-tuning outperforms top-tuning approach (Stanford Cars).\\
	The training procedure is analogous to the one performed for DenseNet201 with 5 additional pre-trained models and a subset of 4 datasets, performing 600 additional training processes.\\
	In \autoref{fig:Models_accs_times} we show the overall results. Each color refers to a different target dataset, each symbol corresponds to a different pre-trained neural network. 

	\begin{table*}[!t]
		\centering
		\setlength{\tabcolsep}{2pt}
		\begin{tabular}{ccccccccc}
			\hline
			& \multicolumn{2}{c}{Caltech-101} & \multicolumn{2}{c}{CIFAR100} & \multicolumn{2}{c}{DTD} & \multicolumn{2}{c}{Stanford Cars} \\
			& $\Delta$Acc$\uparrow$ & SpUp$\uparrow$ & $\Delta$Acc$\uparrow$ & SpUp$\uparrow$ & $\Delta$Acc$\uparrow$ & SpUp$\uparrow$ & $\Delta$Acc$\uparrow$ & SpUp$\uparrow$\\[2pt]
			\hline
			DenseN201 &$+1.10\%$&$94.00 $&$-3.70\%$&$12.1 $&$+2.90\%$&$82.80 $&$-15.7\%$&$76.90 $\\[2pt]
			Eff.NB0 & $-5.60\%$& $109.3 $& $-12.1\%$& $8.70 $& $-2.60\%$& $76.80 $& $-21.9\%$& $73.40 $\\[2pt]
			Inc.R.NV2 & $-2.80\%$& $85.10 $& $-11.1\%$& $10.2 $& $-1.50\%$& $126.3 $& $-18.3\%$& $137.6 $\\[2pt]
			MobileNV2 & $+14.3\%$& $39.40 $& $+0.10\%$& $7.90 $& $+23.5\%$& $89.80 $& $-4.3\%$& $48.10 $\\[2pt]
			ResN152 & $+2.90\%$& $82.60 $& $-5.10\%$& $13.7 $& $+2.90\%$& $150.2 $& $+7.2\%$& $78.90 $\\[2pt]
			Xception & $-4.00\%$& $67.10 $& $-15.4\%$& $10.3 $& $-3.90\%$& $81.00 $& $-22.7\%$& $98.10 $\\
			
			%  		 DenseN201 &$+1.10\%$&$94.00\times$&$-3.70\%$&$12.1\times$&$+2.90\%$&$82.80\times$&$-15.7\%$&$76.90\times$\\
			% 		 Eff.NB0 & $-5.60\%$& $109.3\times$& $-12.1\%$& $8.70\times$& $-2.60\%$& $76.80\times$& $-21.9\%$& $73.40\times$\\
			% 		 Inc.R.NV2 & $-2.80\%$& $85.10\times$& $-11.1\%$& $10.2\times$& $-1.50\%$& $126.3\times$& $-18.3\%$& $137.6\times$\\
			% 		 MobileNV2 & $+14.3\%$& $39.40\times$& $+0.10\%$& $7.90\times$& $+23.5\%$& $89.80\times$& $-4.3\%$& $48.10\times$\\
			% 		 ResN152 & $+2.90\%$& $82.60\times$& $-5.10\%$& $13.7\times$& $+2.90\%$& $150.2\times$& $+7.2\%$& $78.90\times$\\
			% 		 Xception & $-4.00\%$& $67.10\times$& $-15.4\%$& $10.3\times$& $-3.90\%$& $81.00\times$& $-22.7\%$& $98.10\times$\\
			\hline
		\end{tabular}
		\caption{Analysis on different pre-trained models: variation in accuracy ($\Delta$Acc) and speedup (SpUp), between fine-tuning and top-tuning.}
		\label{table:Models_accs_times}
	\end{table*}

	Different pre-trained neural networks show a similar trend to the one obtained by DenseNet201. \autoref{fig:Models_accs_times}(Left) shows that on datasets where the drop between fine-tuning and top-tuning was marginal (e.g. Caltech, DTD) using different pre-trained model results in an analogous behavior. Similarly, on a dataset where such the drop was greater (e.g. Stanford Cars) using different pre-trained results in a similar performance deterioration. \autoref{fig:Models_accs_times}(Right) confirms our findings on the matter of training time speedup, showing that the top-tuning approach is highly faster with respect to the fine-tuning one. \autoref{table:Models_accs_times} summarizes the results obtained both in terms of accuracy and training time speed-up. It shows quantitatively that different pre-trained networks behave similarly on the same dataset with respect to DenseNet201. Moreover, it shows that the top-tuning approach is approximately 70 times faster with respect to the fine-tuning one. Such results suggest that our findings are low dependent on the pre-trained neural network adopted.

	\paragraph{Results with different fine-tuning configurations}
	We further evaluate different protocols of fine-tuning, to investigate the impact of such protocols on training time and test accuracy.
	In the new set of experiments, we explore intermediate fine-tuning configurations, considering our best-performing pre-trained convolutional model, DenseNet201, which consists of 5 dense blocks in its convolutional part. 
	Thus, we compare three partial fine-tuning protocols:
	\begin{itemize}
		\item[-] Updating the last two dense blocks of the convolutional part, which corresponds to updating 47.5\% of the weights in the convolutional part.
		\item[-] Updating the last four dense blocks of the convolutional part, which corresponds to updating 92.1\% of the weights in the convolutional part.
		\item[-] Updating all five dense blocks, which corresponds to updating 100\% of the weights in the convolutional part. It is noteworthy that this scenario corresponds to the original fine-tuning configuration.
	\end{itemize}
	
	\begin{figure*}[!t]
		\centering
		{\includegraphics[width=0.5\textwidth]{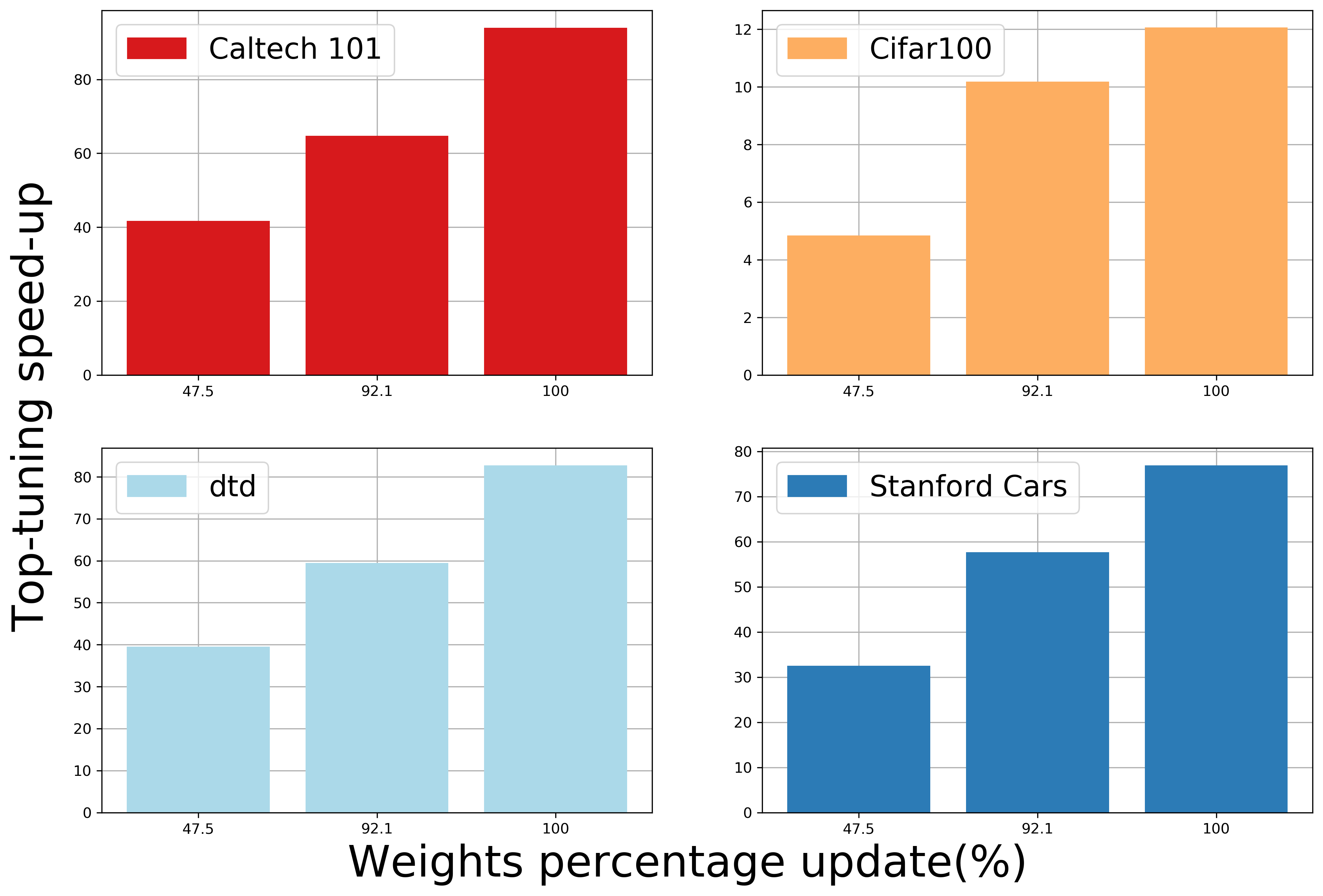}}
		\hspace{3mm}
		{\includegraphics[width=0.44\textwidth]{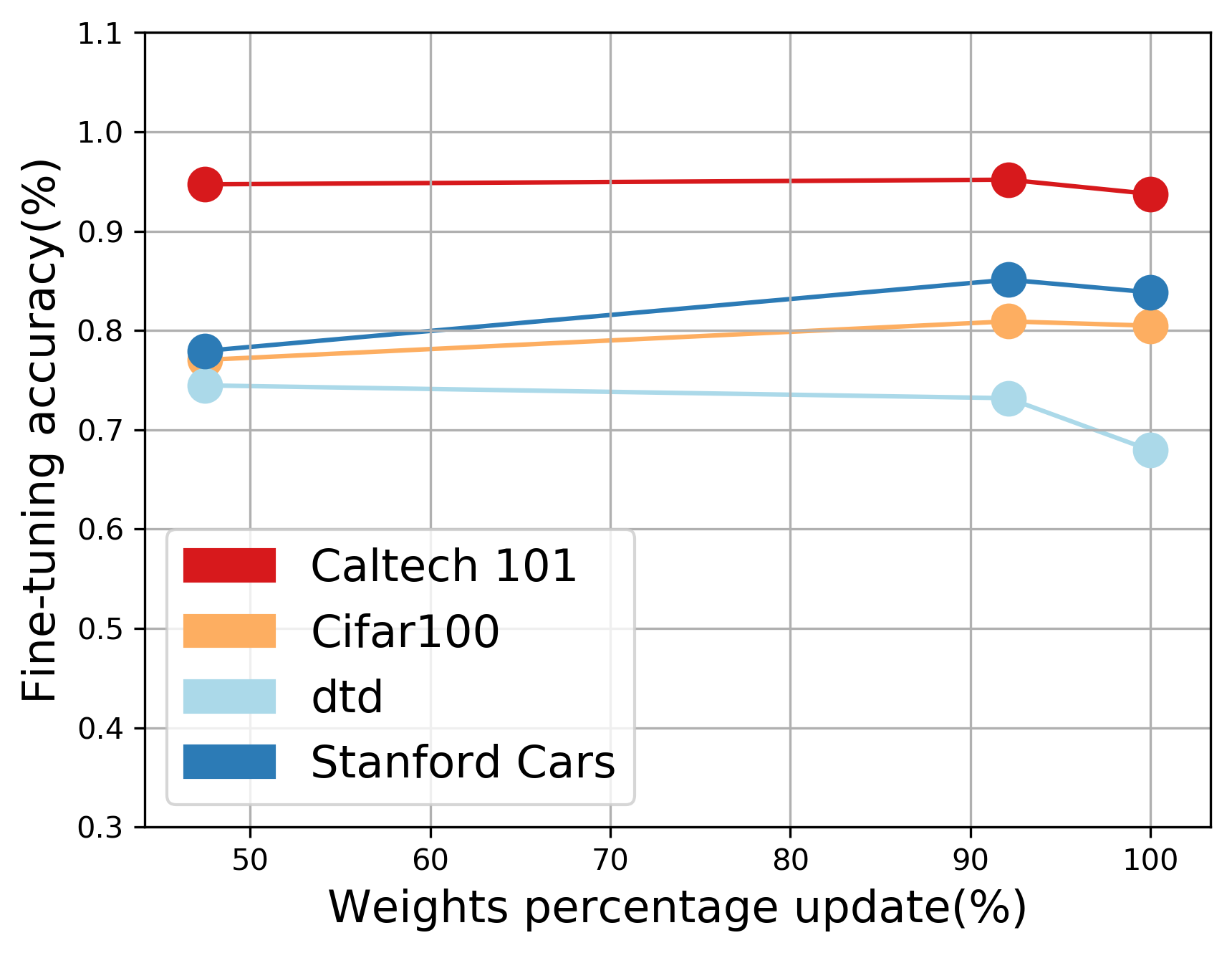}}
		\caption{Overall results on different fine-tuning configurations. (Left) Speed-up obtained over different target datasets by the top-tuning model w.r.t different fine-tuning configurations (Right) Accuracies obtained over different target datasets with different partial fine-tuning configurations.\vspace*{3mm}}
		\label{fig:Freezed_times_accs}
	\end{figure*}

	In \autoref{fig:Freezed_times_accs}(Left), we present the speedup results. We compute the ratio between each intermediate fine-tuning configuration training time and top-tuning training time. As expected, the larger the updated portion of the convolutional part, the greater the cost of fine-tuning, resulting in a higher speed-up of the top-tuning approach.\\
	In \autoref{fig:Freezed_times_accs}(Right), we show the accuracy results for the different fine-tuning configurations. We can notice that, within the same dataset, there is no substantial difference in accuracy among the various configurations. These results seem to confirm that the ImageNet learned representation is very effective, suggesting that the top-tuning approach may be an efficient alternative to fine-tuning for the investigated datasets.

	\paragraph{Results with transformers}
	In the last few years, a new generation of models called \textit{transformers}, has been introduced. Such models are usually composed of hundreds of millions of parameters, obtaining state-of-the-art performances. Commonly, they are pre-trained on an extended version of ImageNet called ImageNet21k\cite{DBLP:journals/corr/abs-2104-10972}. In the following experiments we replace the first convolutional part of the pipeline with a Vision Transformer(ViT-L/16) as presented in \cite{vit2020}. With 32 target datasets for the top-tuning approach and 2 target datasets for the fine-tuning one, we perform $650$ additional training processes.\\
	For the top-tuning pipeline, we were able to replicate our analysis on all 32 target datasets. We report the results in \autoref{table:Models_accs_transf}, where $\Delta Acc= Acc_{\mathrm{Transform}} - Acc_{\mathrm{Conv}}$ is the accuracy gain of transformer with respect to  DenseNet201. On average $\Delta Acc=4.58\% \pm 5.36\%$ shows good improvements by using the pre-trained transformer as a features extractor.
	
	\begin{table*}[!t]\setlength{\tabcolsep}{3pt}
		\centering
		\begin{tabular}{cc}
			\hline
			Dataset& $\Delta$Acc$\uparrow$\\[2pt]
			\hline
			AFHQ & $-0.03\%$\\[2pt]
			Beans & $3.91\%$ \\[2pt]
			Best art & $10.1\%$\\ [2pt]
			Boat & $7.89\%$\\[2pt]
			Cal101 &  $1.78\%$\\[2pt]
			Cassava &  $8.07\%$\\[2pt]
			Cat Do &  $1.14\%$\\[2pt]
			Ch xra &  $-0.65\%$\\[2pt]
			\hline
		\end{tabular}
		\begin{tabular}{cc}
			\hline
			Dataset& $\Delta$Acc$\uparrow$\\[2pt]
			\hline
			CIF10 & $5.92\%$\\[2pt]
			CIF100 &  $14.56\%$\\[2pt]
			Citr lea &  $2.67\%$\\[2pt]
			Col hist & $0.72\%$\\ [2pt]
			Deep w & $11.1\%$\\ [2pt]
			DTD &  $9.98\%$\\[2pt]
			EuSAT & $0.87\%$\\[2pt]
			FG Air & $-9.40\%$\\[2pt]
			\hline
		\end{tabular}
		\begin{tabular}{cc}
			\hline
			Dataset& $\Delta$Acc$\uparrow$\\[2pt]
			\hline
			Foot Ru & $8.98\%$\\[2pt]
			Gemst &  $6.08\%$\\[2pt]
			Hor Hu &  $0.00\%$\\[2pt]
			iCub & $0.12\%$\\ [2pt]
			Ind Fd & $13.0\%$\\ [2pt]
			MkNoM &  $7.02\%$\\[2pt]
			Malaria & $1.84\%$\\[2pt]
			Meat qu & $-0.63\%$\\[2pt]
			\hline
		\end{tabular}
		\begin{tabular}{cc}
			\hline
			Dataset& $\Delta$Acc$\uparrow$\\[2pt]
			\hline
			Oxf Flo & $12.0\%$\\[2pt]
			Oxf Pet &  $5.75\%$\\[2pt]
			Plankt &  $0.00\%$\\[2pt]
			Sa Cov & $-0.96\%$\\ [2pt]
			St Cars & $4.72\%$\\ [2pt]
			St Dog &  $13.7\%$\\[2pt]
			Ten Fl & $6.98\%$\\[2pt]
			Weath & $-0.53\%$\\[2pt]
			\hline
		\end{tabular}
		\caption{Analysis of the transformer model for the top-tuning pipeline. $\Delta Acc= Acc_{\mathrm{Transform}} - Acc_{\mathrm{Conv}}$ is the accuracy gain of transformer model with respect to convolutional model}
		\label{table:Models_accs_transf}
	\end{table*}

	For the fine-tuning approach, we decided not to replicate the analysis on all the datasets. Indeed, the duration of a single fine-tuning training process can extend for several hours, even on very small datasets. Therefore, to test our findings, we carried out the whole analysis on two datasets of modest size for the sake of comparison. We considered Citrus Leaves and Oxford Flowers, where we reached an absolute accuracy of $97.5\%$ and $99.1\%$, respectively. If we compare these values with the ones obtained by the top-tuning pipeline ($98.3\%$ and $99.5\%$, respectively) the results confirm the marginal accuracy benefits of fine-tuning a transformer model. On the training time, we obtain an even more remarkable speedup($226.19\times$ and $185.05\times$, respectively).

	\subsection{Pre-training dataset influence transfer’s effectiveness}\label{para:Analysis on different pre-trains}
	We now explore the dependency of the results on the pre-training source dataset. 
	%Specifically we fix the pre-trained neural network architecture (DenseNet201), and we investigate the benefit of choosing ImageNet in terms of its size (number of images and classes) and image quality.
	To this purpose, we consider three alternative pre-training datasets: 
	\begin{itemize}
		\item \textbf{CIFAR100}: we consider CIFAR100 as a simplified version of ImageNet. With this dataset, we test the impact of a reduced number of images, classes and image size.
		\item \textbf{ImageNet100}: we extract an ImageNet subset with the same number of images and classes of CIFAR100. To make this dataset as similar as possible to CIFAR100 in terms of label semantics: $75\%$ of labels selected from ImageNet are the same in CIFAR100, $15\%$ of them are similar and $10\%$ are different.
		With this dataset, we test the impact of image quality on the obtained results.
		\item \textbf{ImageNet 50k}: we extract an ImageNet subset that contains all the classes in ImageNet with the same number of data points contained in CIFAR100, that is 50k. The obtained dataset has 1000 classes with only 50 points per class.
		With this dataset, we test how the total number of classes affects the obtained results. 
	\end{itemize}
	
	First, we train a DenseNet201 model from scratch on each of the three pre-training datasets. Then, we apply the two investigated pipelines as described in \autoref{ch:Methodology} on five target datasets: one of the most difficult and one of the easiest datasets for both approaches (DTD and CIFAR10, respectively), one where top-tuning approach outperforms fine-tuning (Citrus Leaves) and the opposite case (Deep Weeds). Lastly, we consider the fine-grained Oxford Flowers 102. 
	With five target datasets and three different pre-training, we perform 450 additional training instances.\\
	\begin{figure*}[!t]
		\centering
		{\includegraphics[width=0.47\textwidth]{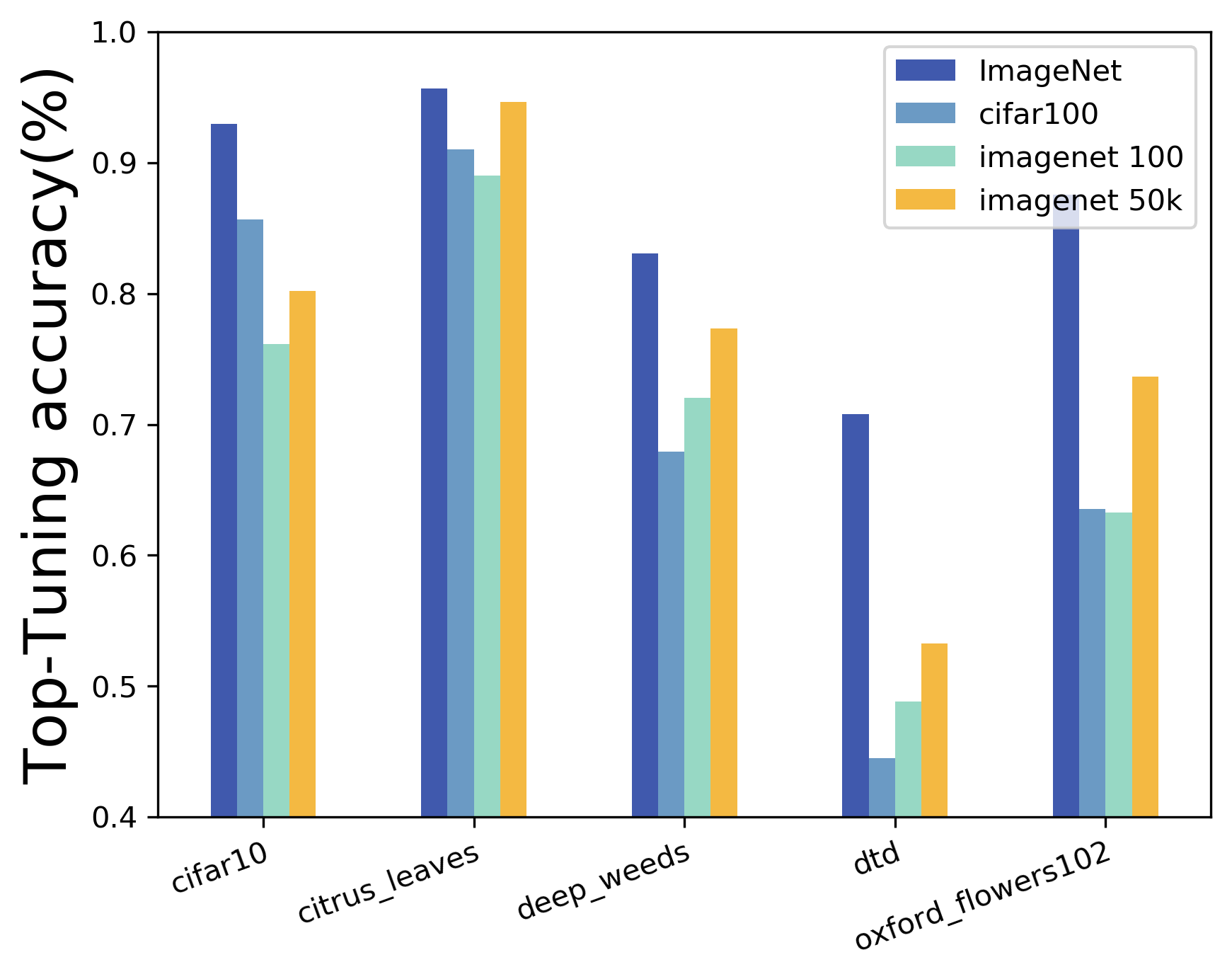}}
		\hspace{3mm}
		{\includegraphics[width=0.47\textwidth]{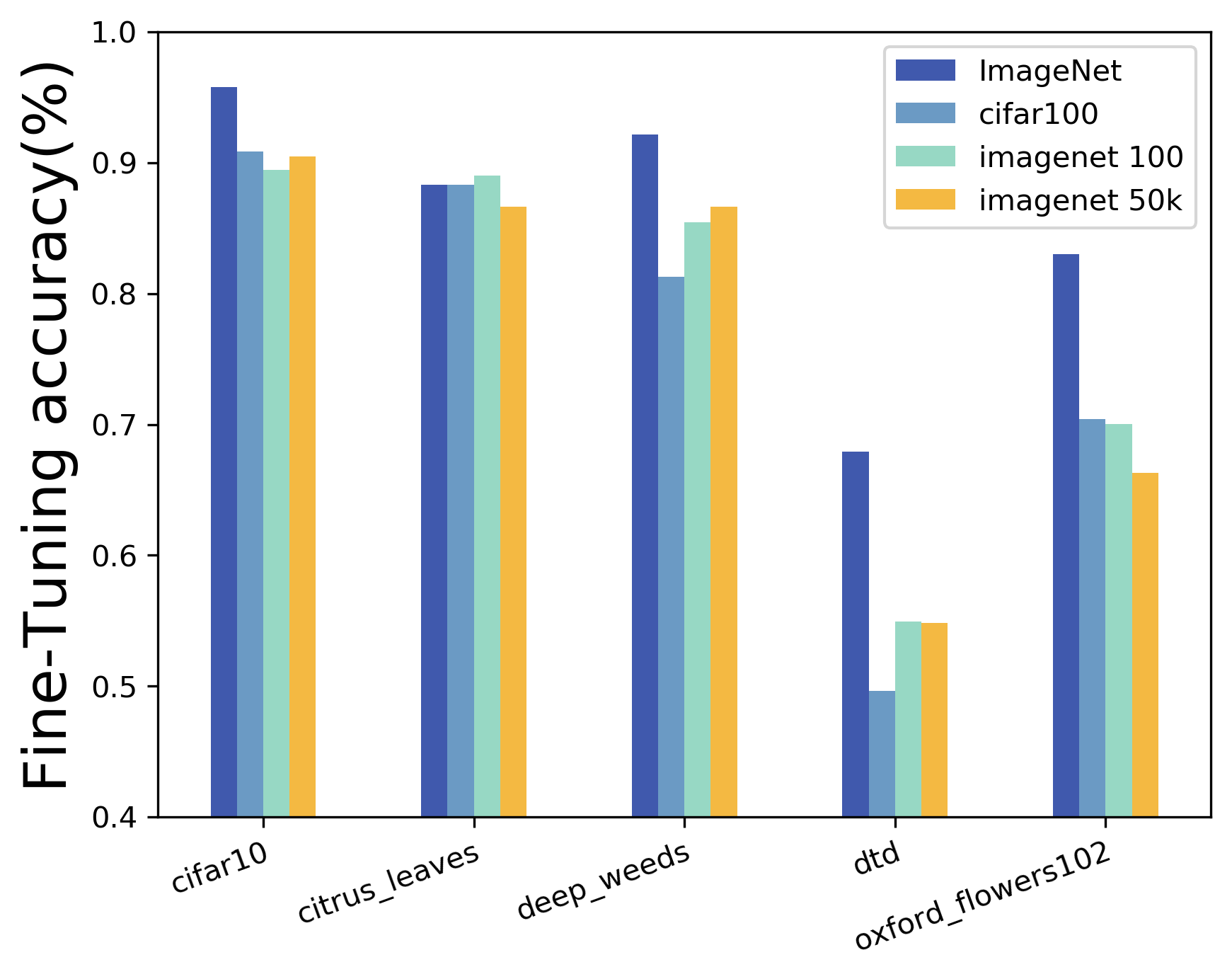}}
		\caption{Overall results on different pre-trains. For each target dataset, each color represents one of the four different pre-trains datasets. (Left) Accuracies obtained by top-tuning approach (Right) Accuracies obtained by fine-tuning approach.}
		\label{fig:Pretrain_accs}
	\end{figure*}
	\begin{table*}[!t]\setlength{\tabcolsep}{3pt}
		\centering
		\begin{tabular}{ccccccc}
			\cline{2-7} & \diagbox{Source}{Target$\downarrow$} &  CIFAR10  &  Citrus Leaves  &  Deep Weeds  &  DTD  &  Oxf Flow \\
			\cline{2-7}
			{\multirow{3}{*}{\rotatebox[origin=c]{90}{\small{Top-Tun.}}}}  & CIFAR100 & $\bm{7.30\%}$& $4.7\%$& $15.2\%$& $26.3\%$& $24.0\%$\\[4pt] 
			& ImageNet100 & $16.8\%$& $6.7\%$& $11.1\%$& $22.0\%$& $24.3\%$\\[4pt] 
			& ImageNet 50k & $12.7\%$& $\bm{1.0\%}$& $\bm{5.70\%}$& $\bm{17.5\%}$& $\bm{13.9\%}$\\

			\cline{2-7}
			{\multirow{3}{*}{\rotatebox[origin=c]{90}{\small{Fine-Tun.}}}} & CIFAR100 & $\bm{4.9\%}$& $0.0\%$& $10.9\%$& $18.3\%$& $\bm{12.6\%}$\\[4pt] 
			& ImageNet100 & $6.3\%$& $\bm{-0.6\%}$& $6.7\%$& $\bm{13.0\%}$& $13.0\%$\\ [4pt]
			& ImageNet 50k & $5.3\%$& $1.7\%$& $\bm{5.5\%}$& $13.1\%$& $16.7\%$\\
			\cline{2-7}
		\end{tabular}
		
		\caption{Accuracy drops for pre-training alternatives to ImageNet. Each column represents a different target dataset, and each row is a different pre-train source. The reported value corresponds to the accuracy drop with respect to the original ImageNet pre-training. The lower the value, the better.}
		\label{table:Pretrains_accs}
	\end{table*} 
	\autoref{fig:Pretrain_accs}(Left) refers to top-tuning accuracy for each pre-train.
	ImageNet pre-training brings the best results on the target datasets. ImageNet 50k is the best alternative, suggesting that the number of classes for the pre-training dataset has a great impact on the top-tuning approach. The only exception corresponds to CIFAR10, where CIFAR100 corresponds to the best pre-train. This behavior depends on CIFAR10 being de facto a CIFAR100 subset. 
	
	\autoref{fig:Pretrain_accs}(Right) refers to fine-tuning accuracy for each pre-train. ImageNet pre-training corresponds to the best accuracy. The difference between the other pre-training configurations is marginal.
	This is probably due to the fine-tuning training procedure, which mitigates the weight of different pre-training.
	\autoref{table:Pretrains_accs} summarizes the quantitative results in terms of accuracy drop with respect to the original ImageNet pre-training.

	\section{Discussion}\label{ch:Discussion}
	In the previous section, we present results describing the efficiency of an approach named \textit{top-tuning}, for image classification tasks. We show that, with respect to conventional fine-tuning procedure, the top-tuning method is between 10 and 165 times faster to be trained, with either an absent or marginal accuracy drop. \\The results obtained in this study pave the way for further developments.\\
	Primarily, it is worth noticing that all the presented results exclusively pertain to the resolution of image classification problems. This task, although important, represents only a part of the topics addressed within the computer vision research field. An interesting development could evaluate the aforementioned techniques' applicability to tasks requiring more structural understanding, such as detection or segmentation. Another potential extension could explore the effectiveness of top-tuning on video-related problems. Indeed, from this perspective, we are currently working on extending the top-tuning pipeline to the field of video action recognition, where efficiency is vital due to the large amount of data involved in video analysis. An additional extension could concern the automatic hyperparameters setting within the top-tuning method. We perform experiments exploiting an automatic hyper-parameter optimization procedure. The resulting accuracy is generally lower with respect to the grid search adopted in this work. 
	This is probably due to the high dimensionality of the data compared to the usual regime for fast-kernel methods optimization, which is typically low-dimensional, leaving room for potential developments. In this regard, we are working to enhance the efficacy of the automatic procedure, with the aim of making the use of the top-tuning approach even more effective.\\
	Secondly, we identify conditions where the top-tuning in its present formulation is less effective in terms of accuracy. 
	This occurs when the target dataset simultaneously exhibits the following characteristics: it is underrepresented in the source dataset, it is of small size, and it is fine-grained. However, based on our experiments, this situation occurs rarely, as the aforementioned criteria are quite strong to satisfy at the same time. Moreover, despite a major accuracy drop, the top-tuning approach is still massively faster which can be the most crucial aspect. Finally, it is worth noting that even the fine-tuning procedure for fine-grained cases may lead to suboptimal results, as handling fine-grained data may require the use of appropriate strategies aimed at highlighting minimal differences between distinct classes.
	
	\section{Conclusions}\label{ch:Conclusions}
	The need for efficiency has become central in the context of deep learning, where complex models are trained with increasingly large datasets. In this work, we compare two popular transfer learning solutions in terms of training time and accuracy. The first one consists in fine-tuning a model pre-trained on a large source dataset. This approach may require significant computational resources, in terms of training time, GPU-CPU involvement, and memory usage. This is due to back-propagation and the potentially huge amount of parameters involved. An alternative solution 
	%often dismissed by the research community as "too naif",
	consists in adopting a pre-trained model as-it-is as a features extractor, coupling it with an external head classifier, which we refer to as {\em top-tuning}\\
	In this paper, we discuss the benefits of this alternative, in particular when a fast kernel head classifier is adopted. To support our claim, we reported an extensive experimental analysis involving 32 small to medium-sized target datasets through 3460 distinct training processes. Our experiments confirm that the top-tuning approach leads to a huge reduction in terms of training time, even from hours to a few minutes in different scenarios. Furthermore, fine-tuning has only marginal accuracy benefits with respect to the top-tuning approach. Indeed, in 60$\%$ of our experiments, $\Delta$Acc between fine-tuning  and top-tuning is in range  $[ -2.5\% ,+2.5\%]$. Moreover, our results showed that the marginal benefit of fine-tuning is low dependent on the neural network architecture used as a pre-trained model.\\
	Another noteworthy aspect of this research is the demonstration of the efficacy of the Nystr{\"o}m approximated kernel method, which the top-tuning model relies on. Indeed, when combined with meaningful pre-trained features, we show its wide applicability to the domain of image classification.\\
	Finally, by considering alternative pre-training datasets, in this paper we emphasize the importance of semantic diversity, showing that greater variability in the source data can have a significant impact on the target task.
	\section*{Acknowledgements}
	
	P.D.A. and L.R. acknowledge the financial support of the European Research Council (grant SLING 819789). V.P.P. acknowledges fundings from FSE REACT-EU-PON 2014–2020, DM 1062/2021

	\bibliography{DLJbib}

\end{document}